\documentclass[10pt,twocolumn,letterpaper]{article}
\usepackage{comment}
\usepackage{cvpr}              

\usepackage{graphicx}
\usepackage{amsmath}
\usepackage[square,sort,comma,numbers]{natbib}
\usepackage{cancel}
\usepackage{amssymb}
\usepackage{booktabs}
\usepackage{bm}
\usepackage{multirow,helvet,booktabs,courier,bm,cite}
\usepackage{amsthm}
\usepackage{times}  
\usepackage{helvet}  
\usepackage[T1]{fontenc}
\usepackage{comment}
\usepackage{diagbox}
\usepackage{courier}  
\usepackage[hyphens]{url}  
\usepackage{graphicx} 
\usepackage{natbib}  
\usepackage{caption} 
\usepackage{algorithm}
\usepackage{algorithmic}

\usepackage[pagebackref,breaklinks,colorlinks]{hyperref}

\usepackage[capitalize]{cleveref}
\crefname{section}{Sec.}{Secs.}
\Crefname{section}{Section}{Sections}
\Crefname{table}{Table}{Tables}
\crefname{table}{Tab.}{Tabs.}

\def\cvprPaperID{438} 
\def\confName{CVPR}
\def\confYear{2023}

\begin{document}
	
	\title{Uncertainty-Aware Unsupervised  Image Deblurring with Deep Residual Prior}
	
	\author{Xiaole Tang$^1$~~~~Xile Zhao$^{1*}$~~~~Jun Liu$^{2*}$~~~~Jianli Wang$^1$~~~~Yuchun Miao$^1$~~~~ Tieyong Zeng$^3$\\
		$^1$University of Electronic Science and Technology of China, Chengdu, China\\
		$^2$Northeast Normal University, Changchun, China\\
		$^3$The Chinese University of Hong Kong, Shatin, NT, Hong Kong\\
		{\tt\small \{sherlock315,xlzhao122003,junliucd,wangjianli123,szmyc1\}@163.com,zeng@math.cuhk.edu.hk}
	}
	\maketitle
	
	\begin{abstract}
		Non-blind deblurring methods achieve decent performance under the accurate blur kernel assumption. Since the kernel uncertainty (i.e. kernel error) is inevitable in practice, semi-blind deblurring is suggested to handle it by introducing the prior of the kernel (or induced) error. However, how to design a suitable prior for the kernel (or induced) error remains challenging. Hand-crafted prior, incorporating domain knowledge, generally performs well but may lead to poor performance when kernel (or induced) error is complex. Data-driven prior, which excessively depends on the diversity and abundance of training data, is vulnerable to out-of-distribution blurs and images. To address this challenge, we suggest a dataset-free deep residual prior for the kernel induced error (termed as residual) expressed by a customized untrained deep neural network, which allows us to flexibly adapt to different blurs and images in real scenarios. By organically integrating the respective strengths of deep priors and hand-crafted priors, we propose an unsupervised semi-blind deblurring model which recovers the clear image from the blurry image and inaccurate blur kernel. To tackle the formulated model, an efficient alternating minimization algorithm is developed. Extensive experiments demonstrate the favorable performance of the proposed method as compared to model-driven and data-driven methods in terms of image quality and the robustness to different types of kernel error. 
	\end{abstract}
	\footnotetext{*Corresponding author}
	\section{Introduction}
	Image {blurring} is mainly caused by camera shake \cite{fergus2006removing}, object motion \cite{jia2007single}, and defocus \cite{zhou2011coded}. By assuming {the} blur kernel is shift-invariant, the image blurring can be formulated as the following convolution process: 
	\begin{equation}\label{eq1}
		\boldsymbol{y}=\boldsymbol{k} \otimes \boldsymbol{x}+\boldsymbol{n},
	\end{equation}
	where $\boldsymbol{y}$ and $\boldsymbol{x}$ denote the blurry image and the clear image respectively, $\boldsymbol{k}$ represents the blur kernel, $\boldsymbol{n}$ represents the additive Gaussian noise, and $\otimes$ is the convolution operator. To acquire the clear image from the blurry  one, image deblurring has received considerable research attention and related methods have been developed.
	\begin{figure}[t]
		\centering
		\setlength\tabcolsep{1pt}
		\begin{tabular}{cccc}
			\includegraphics[width=0.24\linewidth]{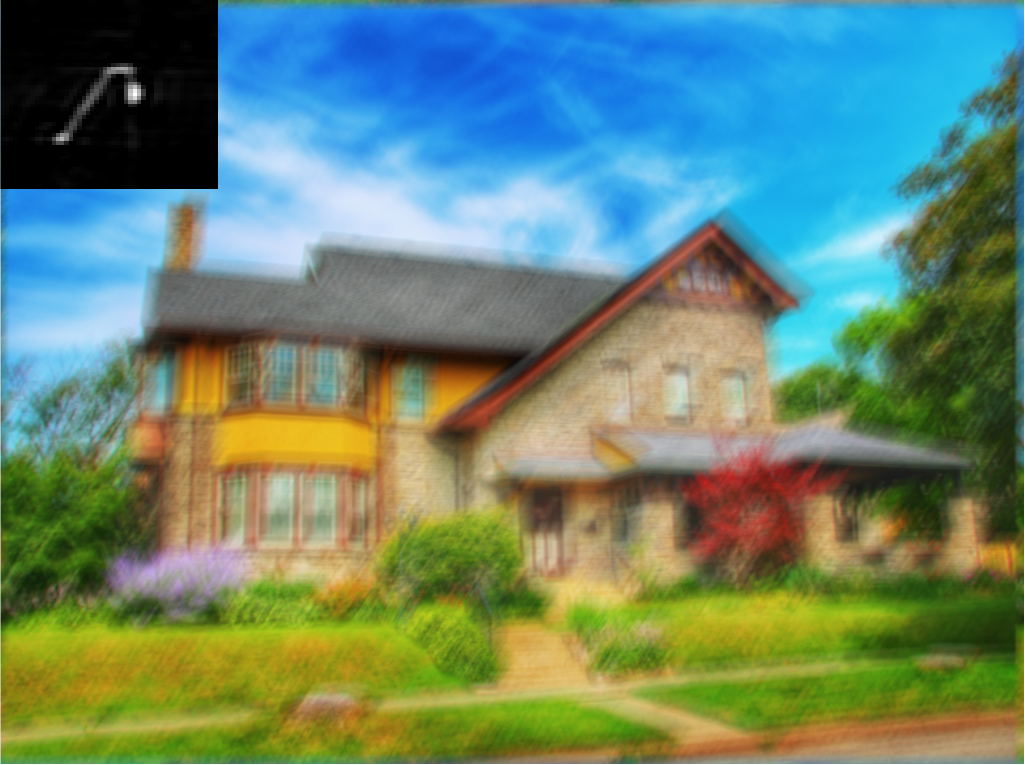}&\includegraphics[width=0.24\linewidth]{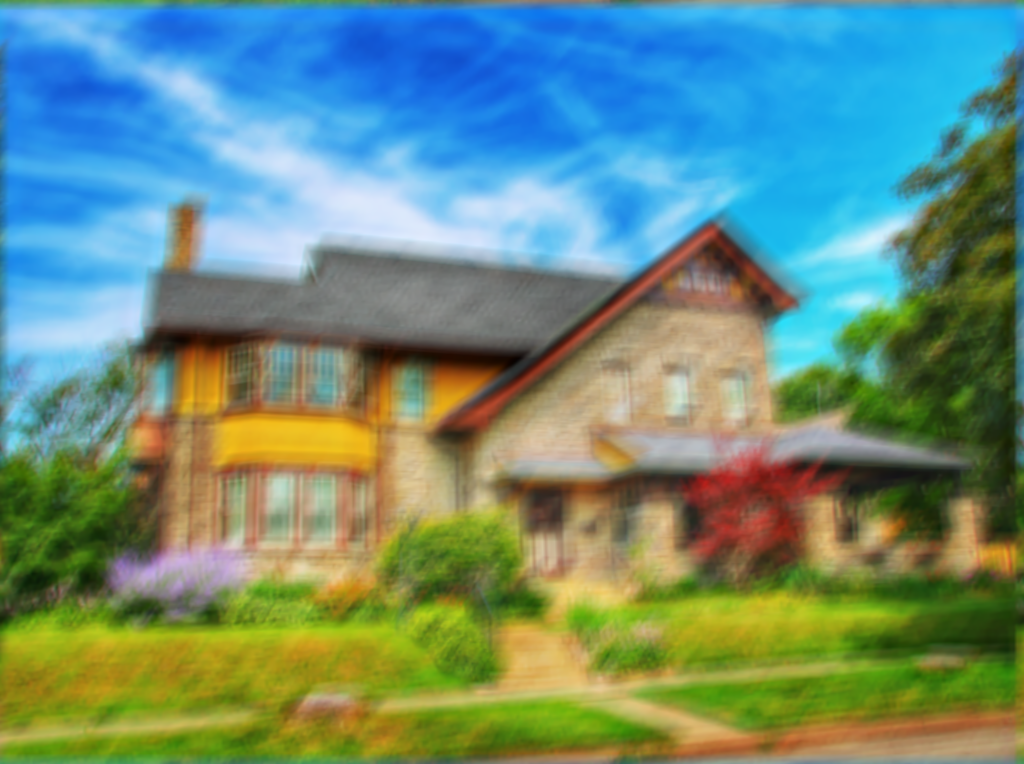}&\includegraphics[width=0.24\linewidth]{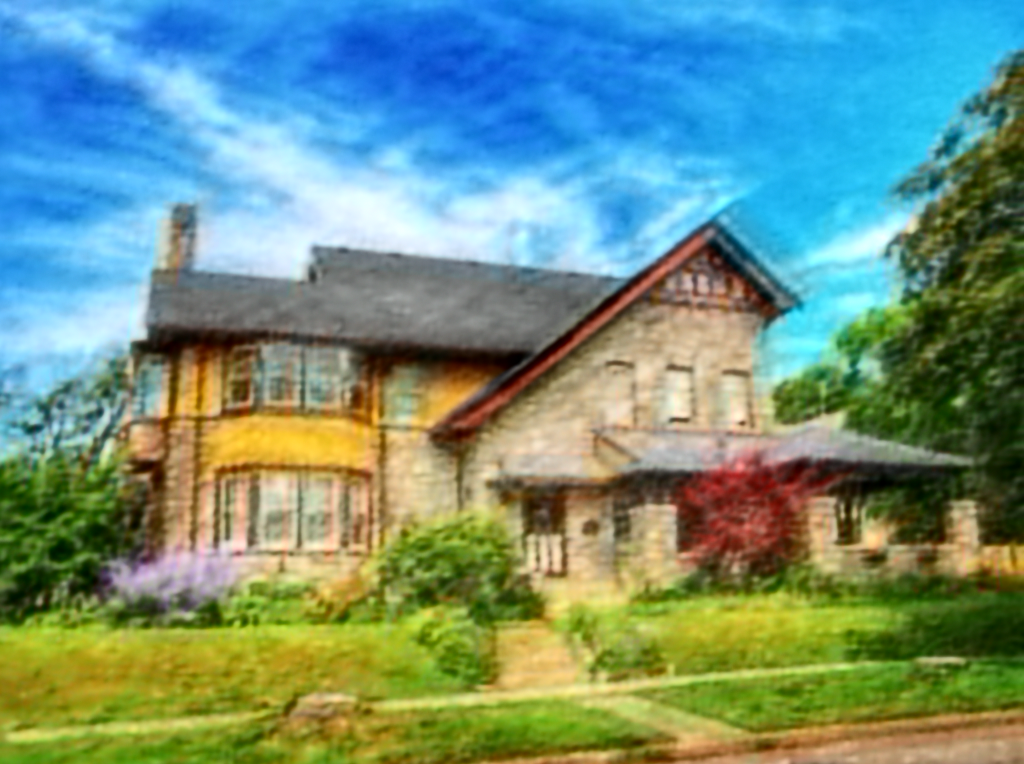}&\includegraphics[width=0.24\linewidth]{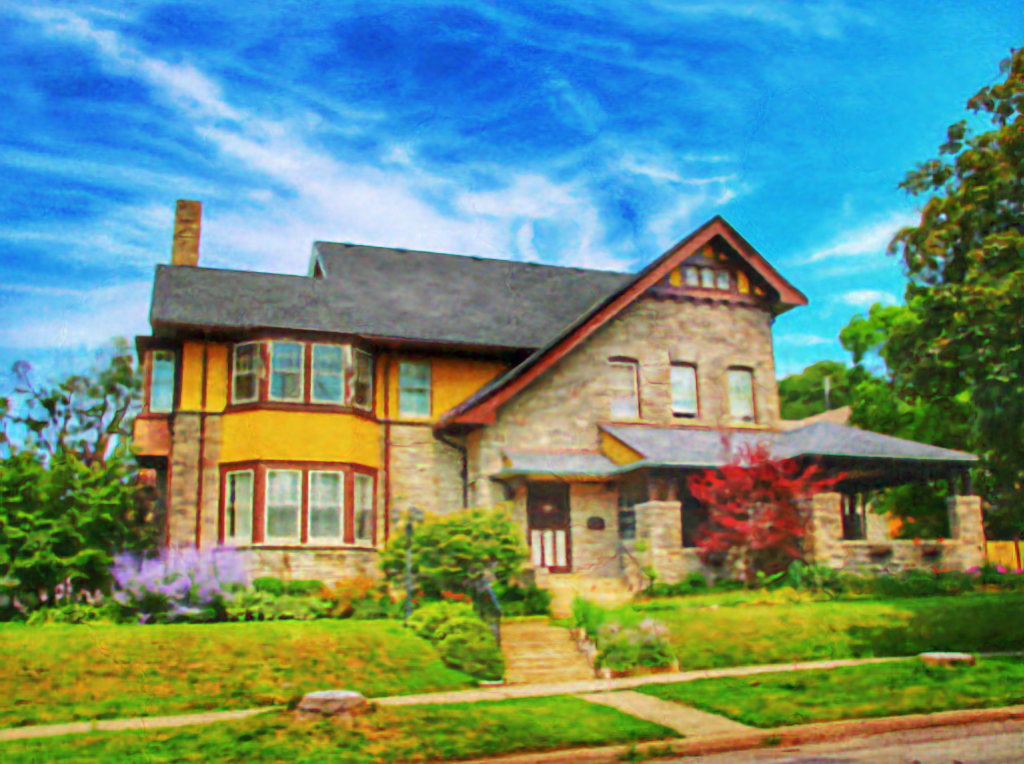}\\
			Blurry&\cite{ji2011robust}&\cite{vasu2018non}&Ours\\
			PSNR 19.56&PSNR 20.83&PSNR 22.75&PSNR \textbf{25.66}\\
			\includegraphics[width=0.24\linewidth]{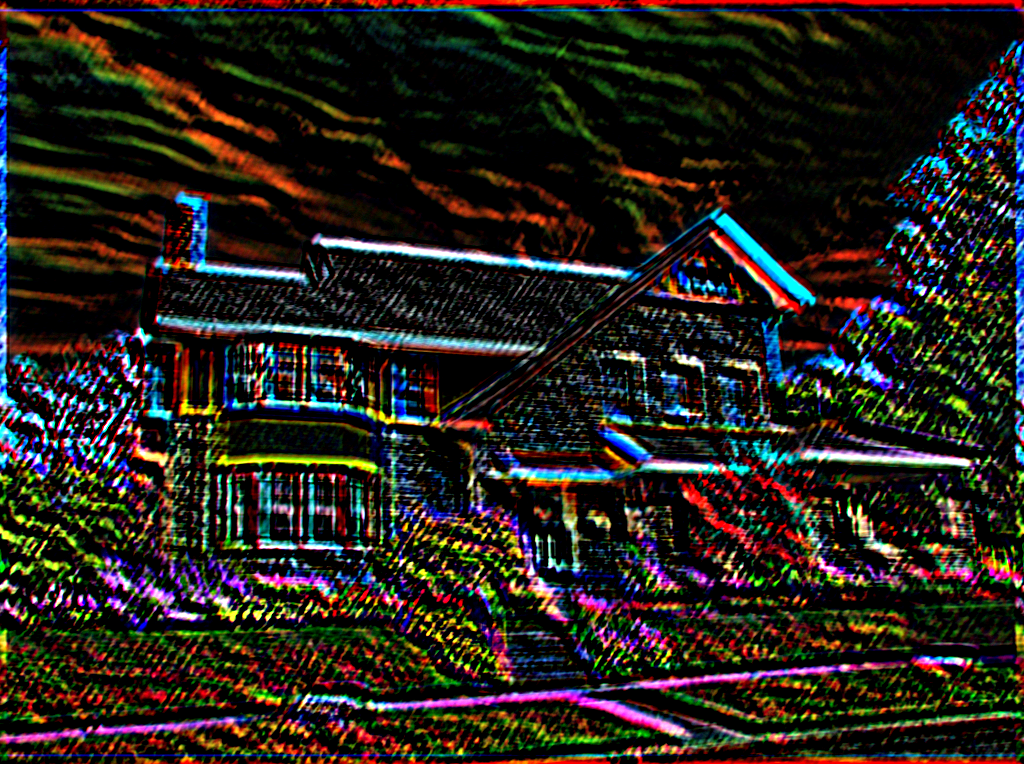}&
			\includegraphics[width=0.24\linewidth]{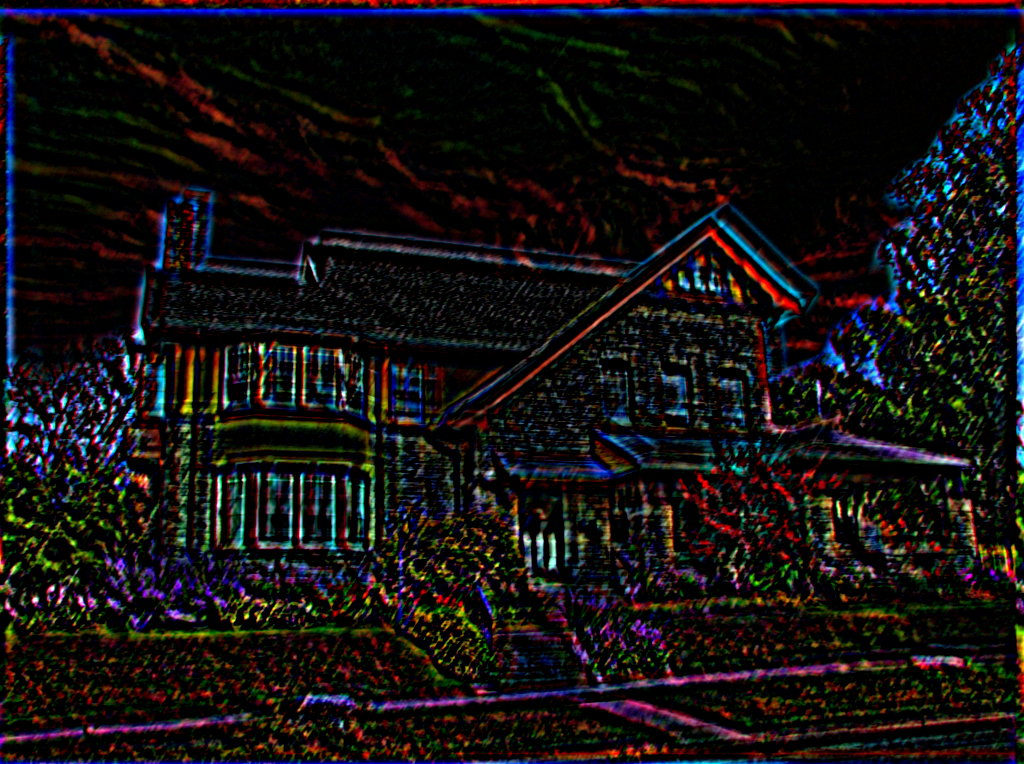}&
			\includegraphics[width=0.24\linewidth]{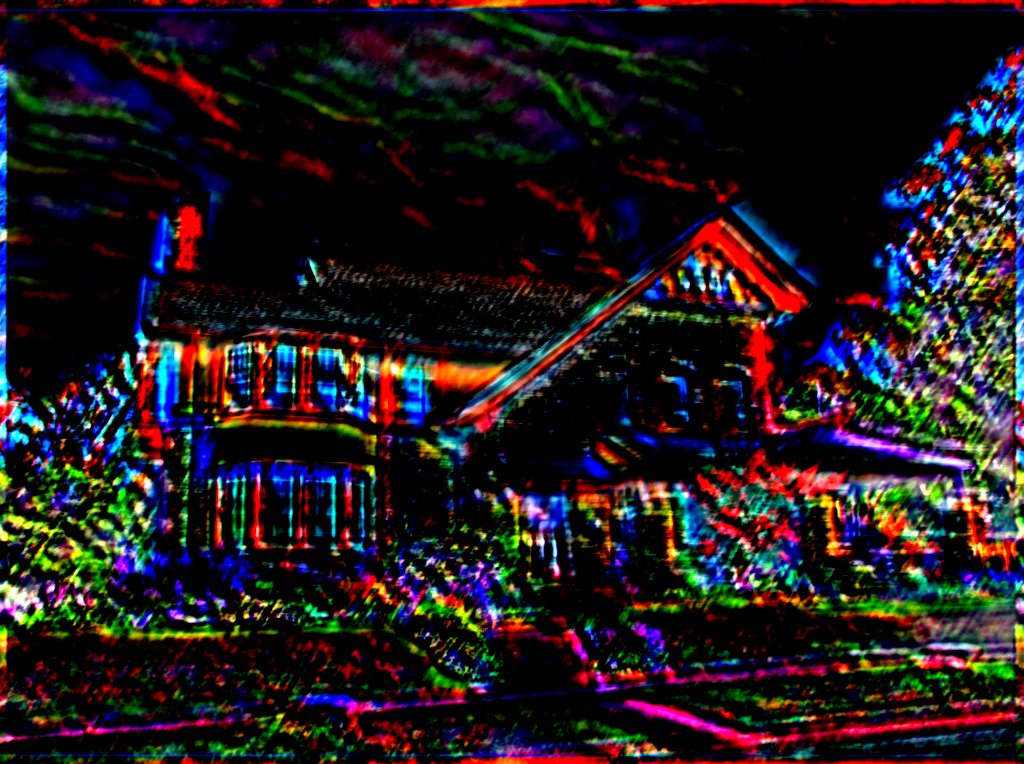}&
			\includegraphics[width=0.24\linewidth]{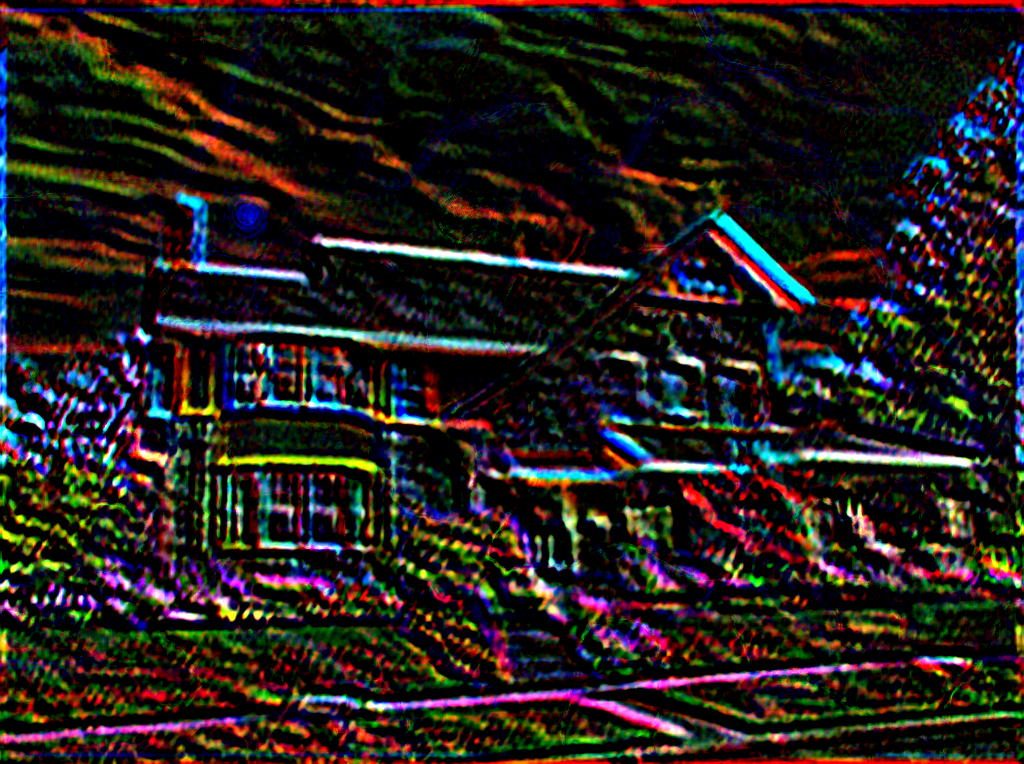}\\
			True Res.&Res. in \cite{ji2011robust}&Res. in \cite{vasu2018non}&Our Res.\\
			MSE 0&MSE 0.075&MSE 0.067&MSE \textbf{0.027}
		\end{tabular}
		\caption{\textbf{Visual comparison of the restored results and estimated residuals} by three semi-blind methods based on different priors for the residual induced by the kernel error, including hand-crafted prior \cite{ji2011robust}, data-driven prior \cite{vasu2018non}, and  the proposed deep residual prior (DRP). The true residual is the convolution result of the kernel error and the clear image ($\bm r=\Delta \boldsymbol{k}\otimes \bm x$). The closer estimated residual is to the true residual, the better it is.}
		\label{1}
	\end{figure}
	
	In terms of the availability of kernel, current image deblurring methods can be mainly classified into two categories, i.e., {\it  blind deblurring methods} in which  the blur kernel is assumed to be unknown, and {\it  non-blind deblurring methods} in which the blur kernel is assumed to be known or computed elsewhere. Typical blind deblurring methods \cite{levin2009understanding, liu2014blind,michaeli2014blind,pan2016blind,ren2016image,sun2013edge,yan2017image, liu2021surface, zuo2016learning, li2022supervised} involve two steps: 1) estimating the blur kernel from the blurry images, and 2) {recovering} the clear image with the estimated blur kernel. Recently there also emerge transformer-based \cite{zamir2022restormer, wang2022general} and unfolding networks \cite{mou2022deep} that learn  direct mappings from blurry image to the deblurred one without using the kernel. Non-blind deblurring methods \cite{wang2008new, krishnan2009fast, 6392274, zhang2017learning1, eboli2020end2end, chen2021learning, dong2021dwdn, quan2021nonblind}, based on various priors for the clear image,  estimate the clear image solely from the blurry image with known blur kernel. Notably, existing non-blind deblurring methods can perform well under the error-free kernel assumption. However, in the real application, uncertainty exists in the kernel acquisition process. As a result, these methods without handling kernel uncertainty often introduce artifacts and cause unpleasant performances.
	
	Recently, semi-blind methods are suggested to handle kernel uncertainty by introducing the prior for the kernel (or induced) error. In the literature, there are two groups of  priors of kernel (or induced) error, i.e., hand-crafted priors and data-driven priors. Hand-crafted priors \cite{ji2011robust, Zhao2013TotalVS}, incorporating domain knowledge, generally perform well but may lead to poor performance when the distribution of kernel (or induced) error is complex. For example, hand-crafted priors (e.g., sparse prior \cite{ji2011robust}) are relatively impotent to characterize the complex intrinsic structure of the kernel induced error; see Figure \ref{1}. Data-driven priors \cite{vasu2018non, nan2020deep, ren2019simultaneous}, which excessively depend on the diversity and abundance of training data, are vulnerable to out-of-distribution blurs and images. Specifically, the data-driven prior in \cite{vasu2018non} that is expressed by a trained network  introduces artifacts around the sharp edges; see Figure \ref{1}. Therefore, how to design a suitable prior for the kernel (or induced) error remains challenging.
	
	To address this problem, we suggest a dataset-free deep prior called deep residual prior (DRP) for the kernel induced error (termed as residual), which leverages the strong representation ability of deep neural networks. Specifically, DRP is expressed by an untrained customized deep neural network. Moreover, by leveraging the general domain knowledge, we use the sparse prior to guide DRP to form a semi-blind deblurring model. This model organically integrates the respective strengths of deep priors and hand-crafted priors to achieve favorable performance. To the best of our knowledge, we are the first to introduce the untrained network to capture the kernel induced error  in semi-blind problems, which is a featured contribution of our work.
	
	In summary,	our contributions are mainly three-fold:
	
	\noindent
	$\bullet$  For the residual induced by the kernel uncertainty, we elaborately design a dataset-free DRP, which allows us to faithfully capture the complex residual in real-world applications as compared to hand-crafted priors and data-driven priors.
	
	\noindent 
	$\bullet$ Empowered by the deep residual prior, we suggest an unsupervised semi-blind deblurring model by synergizing the respective strengths of dataset-free deep prior and hand-crafted prior, which work togerther to deliver promising results.
	
	\noindent
	$\bullet$  Extensive experiments on different blurs and images sustain the favorable performance of our method, especially for the robustness to kernel error.
	
	\section{Related Work}
	This section briefly introduces literatures on semi-blind deblurring methods that focus on handling the kernel error. 
	
	In the early literature of semi-blind deblurring, model-driven methods are dominating, in which these methods generally handle the kernel  uncertainty with some hand-crafted priors. For example,
	Zhao \etal~\cite{Zhao2013TotalVS} directly treated the kernel error as additive zero-mean white Gaussian  noise (AWGN), and $l_2$-regularizer is utilized to model it.  Ji and Wang~\cite{ji2011robust} treated the residual as an additional variable to be estimated, and  a sparse constraint in the spatial domain is utilized to regularize it (i.e., $l_1$-regularizer). However, such hand-crafted priors, which are based on certain statistical distribution assumptions (such as sparsity and AWGN), are not sufficient to characterize the complex structure of kernel error. Thus, such methods are efficient sometimes but not flexible enough to handle the kernel error in complex real scenarios. 
	
	Recently, triggered by the expressing power of neural networks, many deep learning-based methods \cite{ren2017partial, sun2015learning, vasu2018non, nan2020deep, ren2019simultaneous} have emerged, in which data-driven priors are learned from a large number of external data for handling kernel error. The key of these methods is to design an  appropriate neural network structure and learning pipeline. For instance, Ren \etal~ \cite{ren2017partial} suggested a partial deconvolution model to explicitly model kernel estimation error in the Fourier domain. Vasu \etal \cite{vasu2018non} trained their convolutional neural networks with a large number of real and synthetic noisy blur kernels to achieve good performance for image deblurring with inaccurate kernels. Ren \etal~\cite{ren2019simultaneous} suggested simultaneously learning the data and regularization term to tackle the image restoration task with an inaccurate degradation model. Nan and Ji \cite{nan2020deep} cleverly unrolled an iterative total-least-squares estimator in which the residual prior is learned by a customized dual-path U-Net. This method well handles kernel uncertainty  and achieves remarkable deblurring performance in a supervised manner. Although these methods achieve decent performances in certain cases, they excessively depend on the diversity and abundance of training data and are vulnerable to out-of-distribution blurs and images.
	\begin{figure*}
		\centering
		\setlength\tabcolsep{1pt}
		\begin{tabular}{cccc}
			\includegraphics[width=0.24\linewidth]{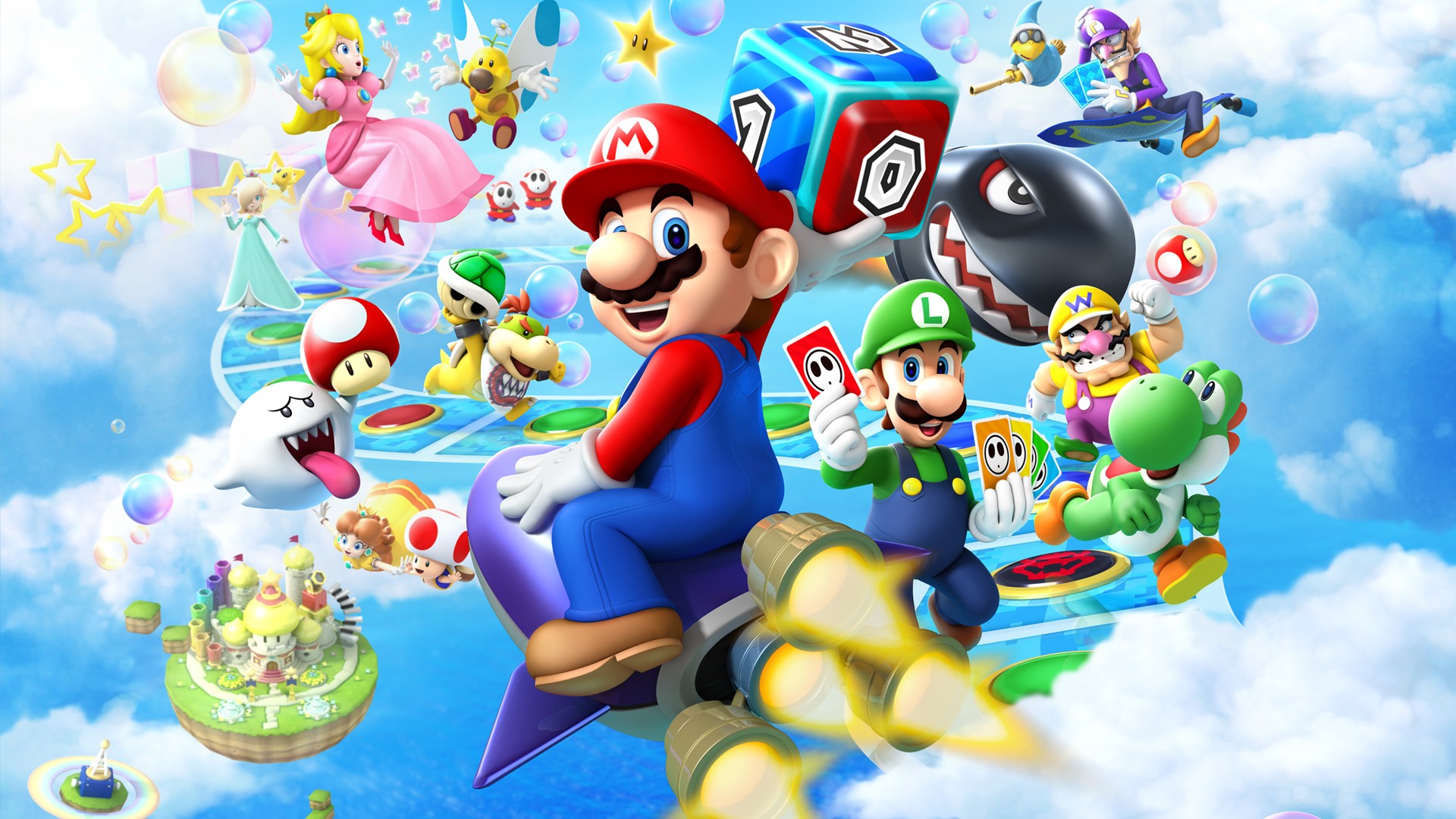}&\includegraphics[width=0.24\linewidth]{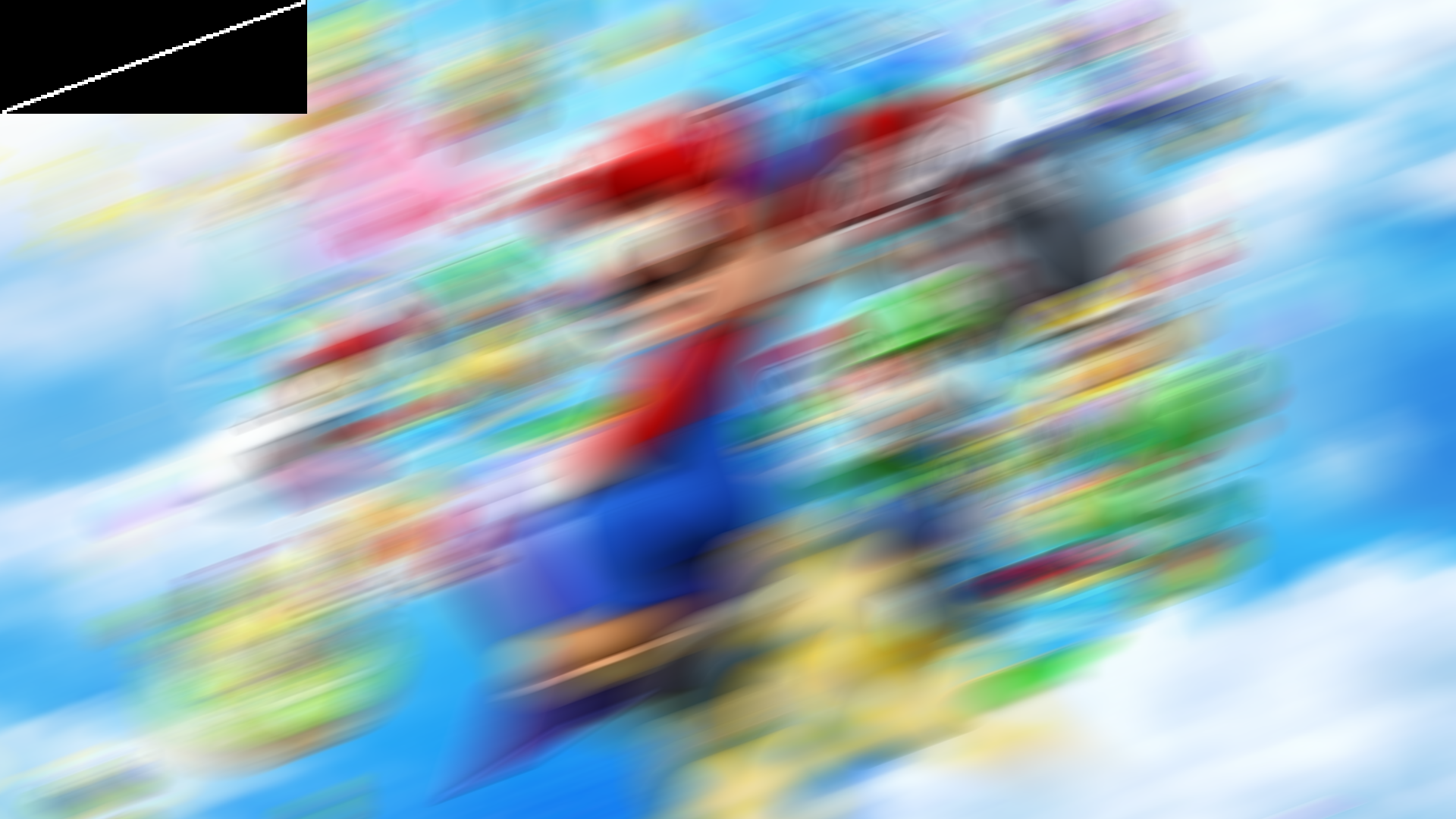}&\includegraphics[width=0.24\linewidth]{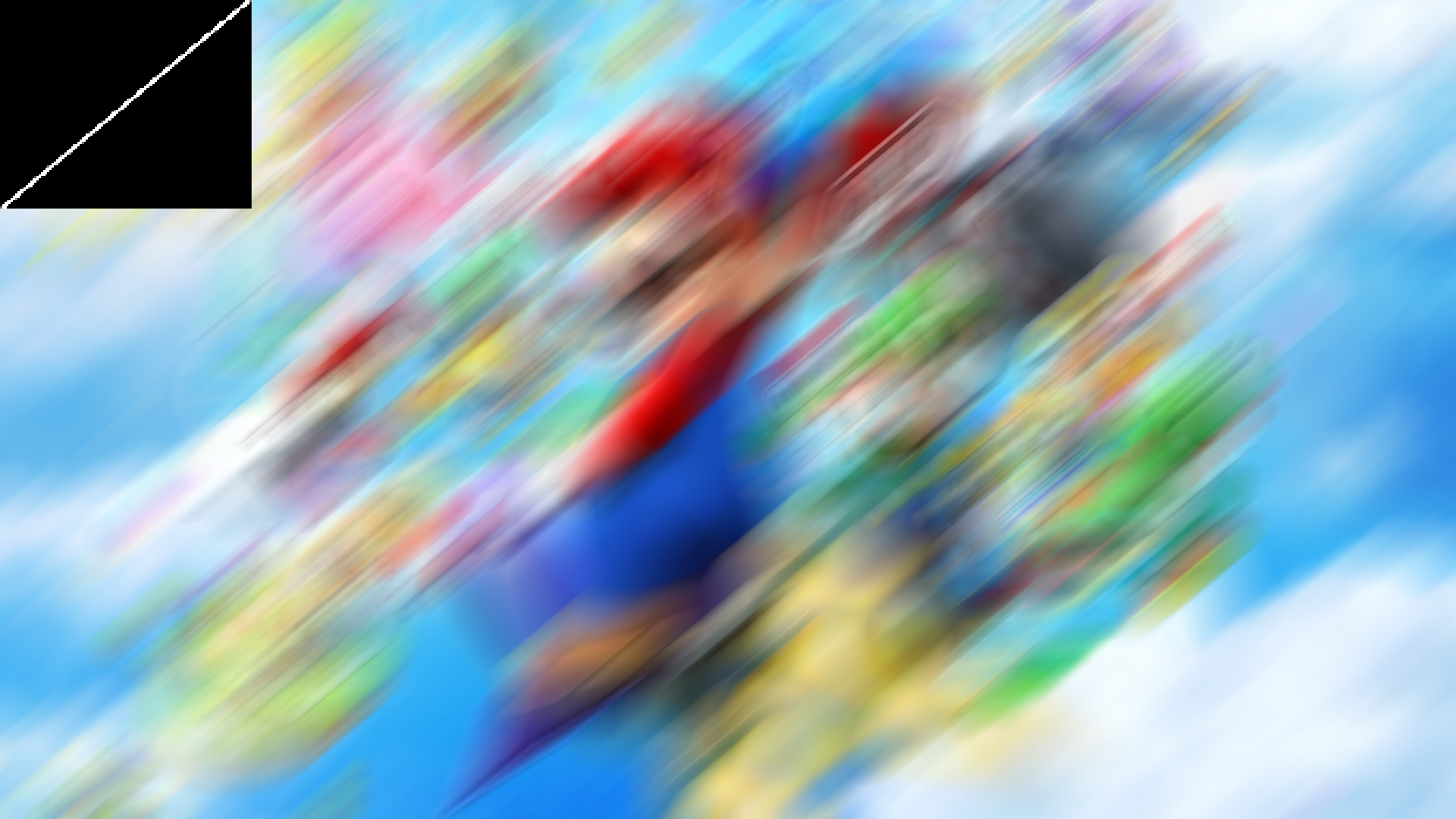}&\includegraphics[width=0.24\linewidth]{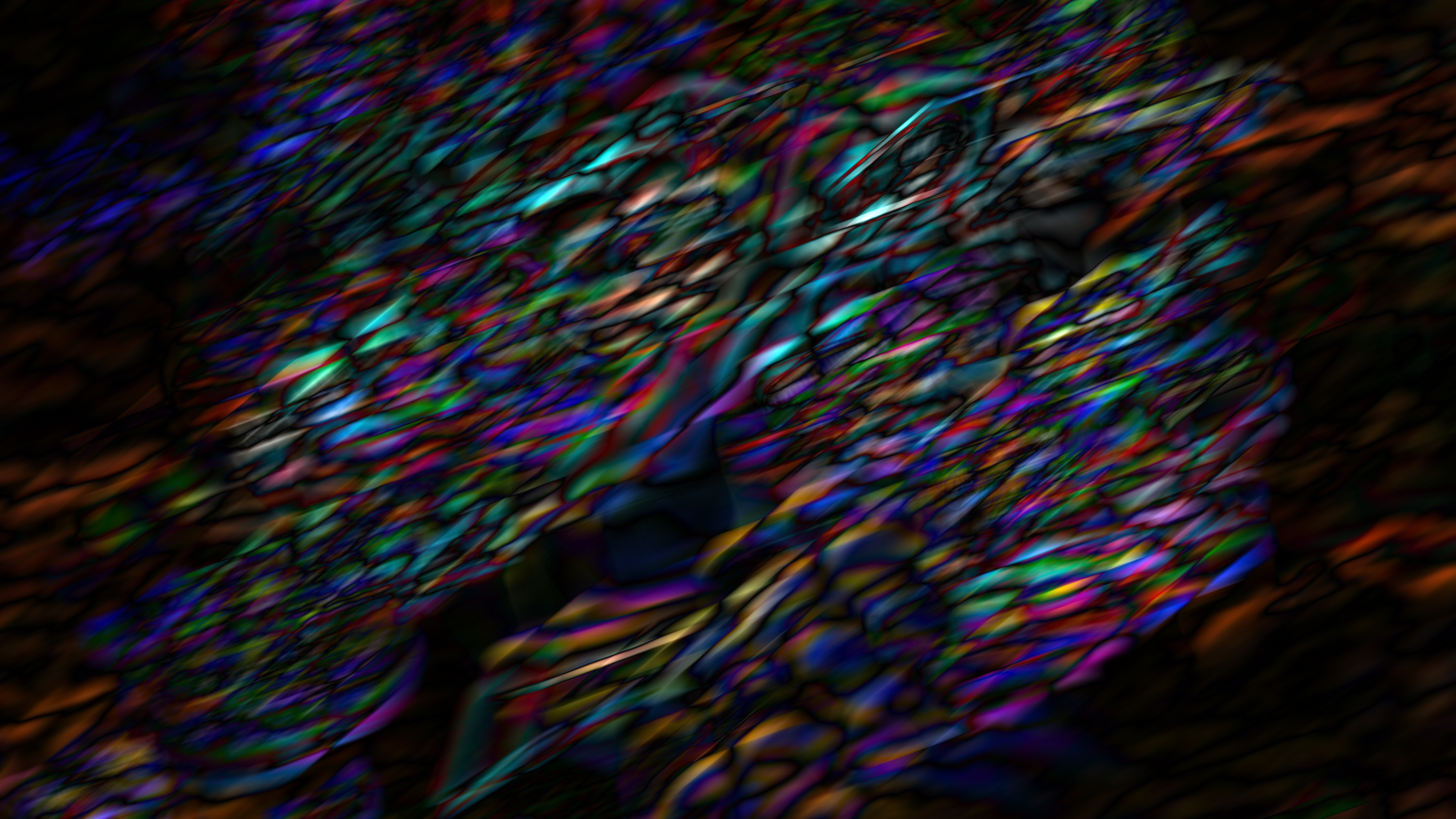}\\
			\includegraphics[width=0.24\linewidth]{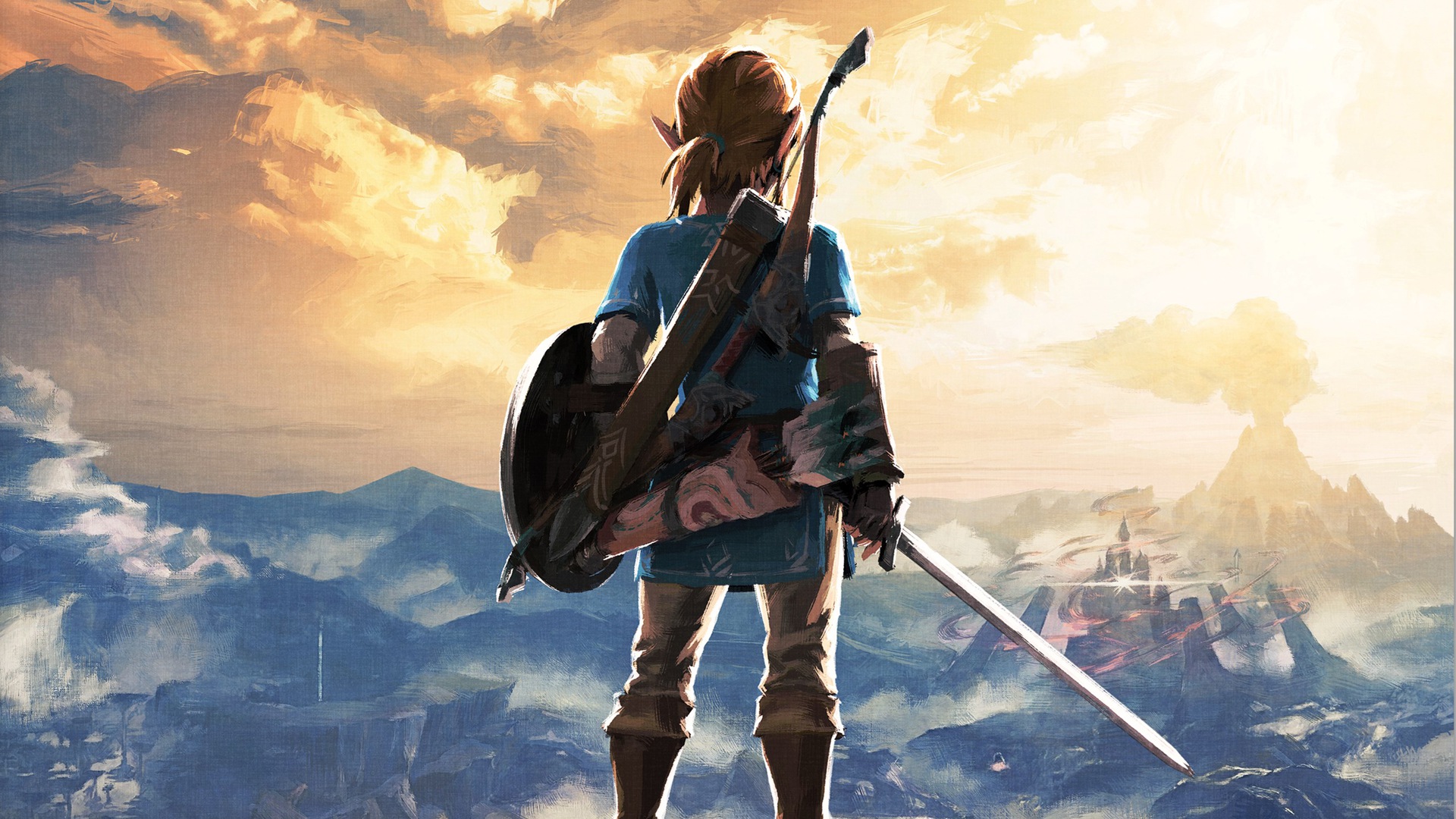}&\includegraphics[width=0.24\linewidth]{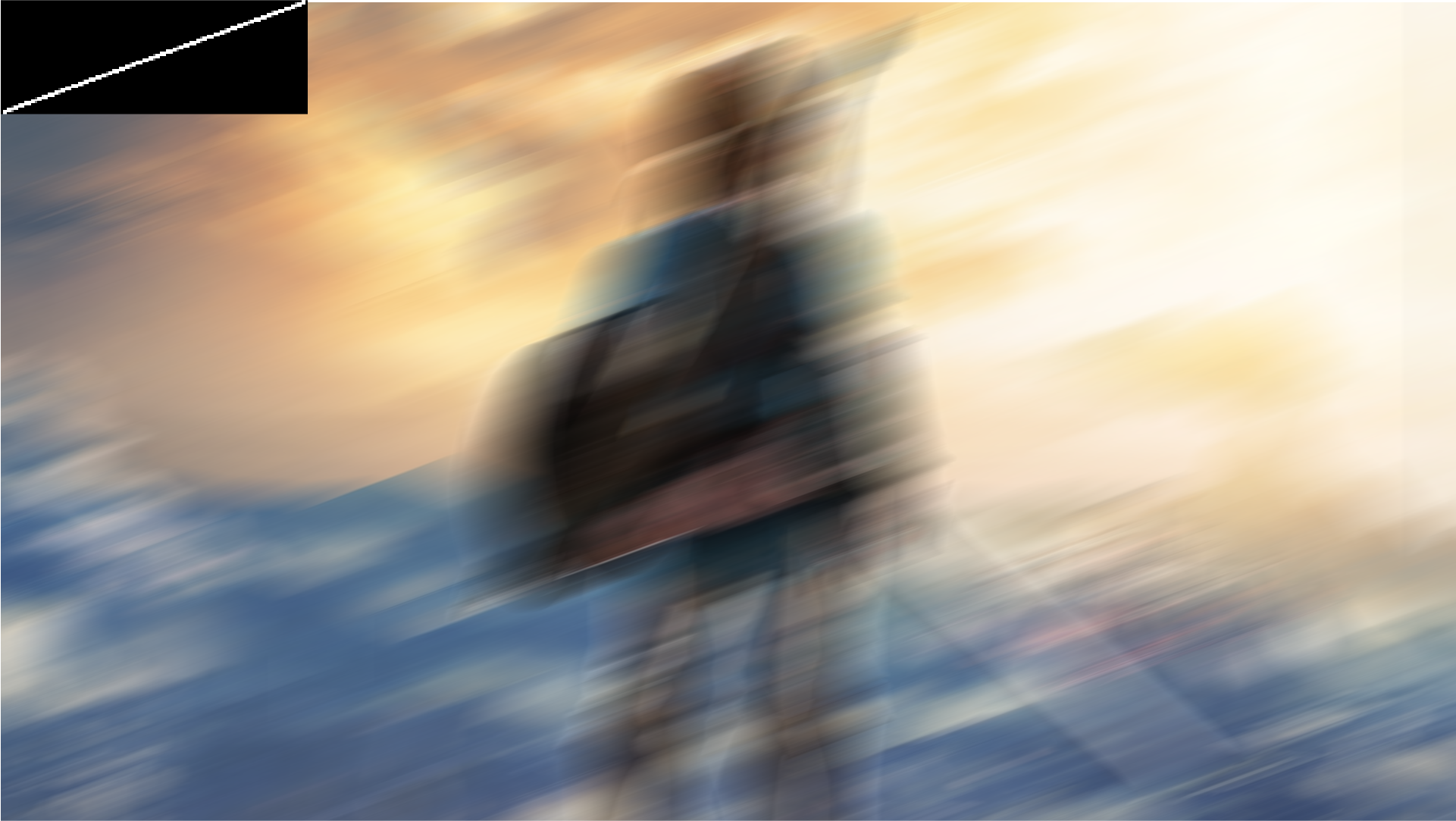}&\includegraphics[width=0.24\linewidth]{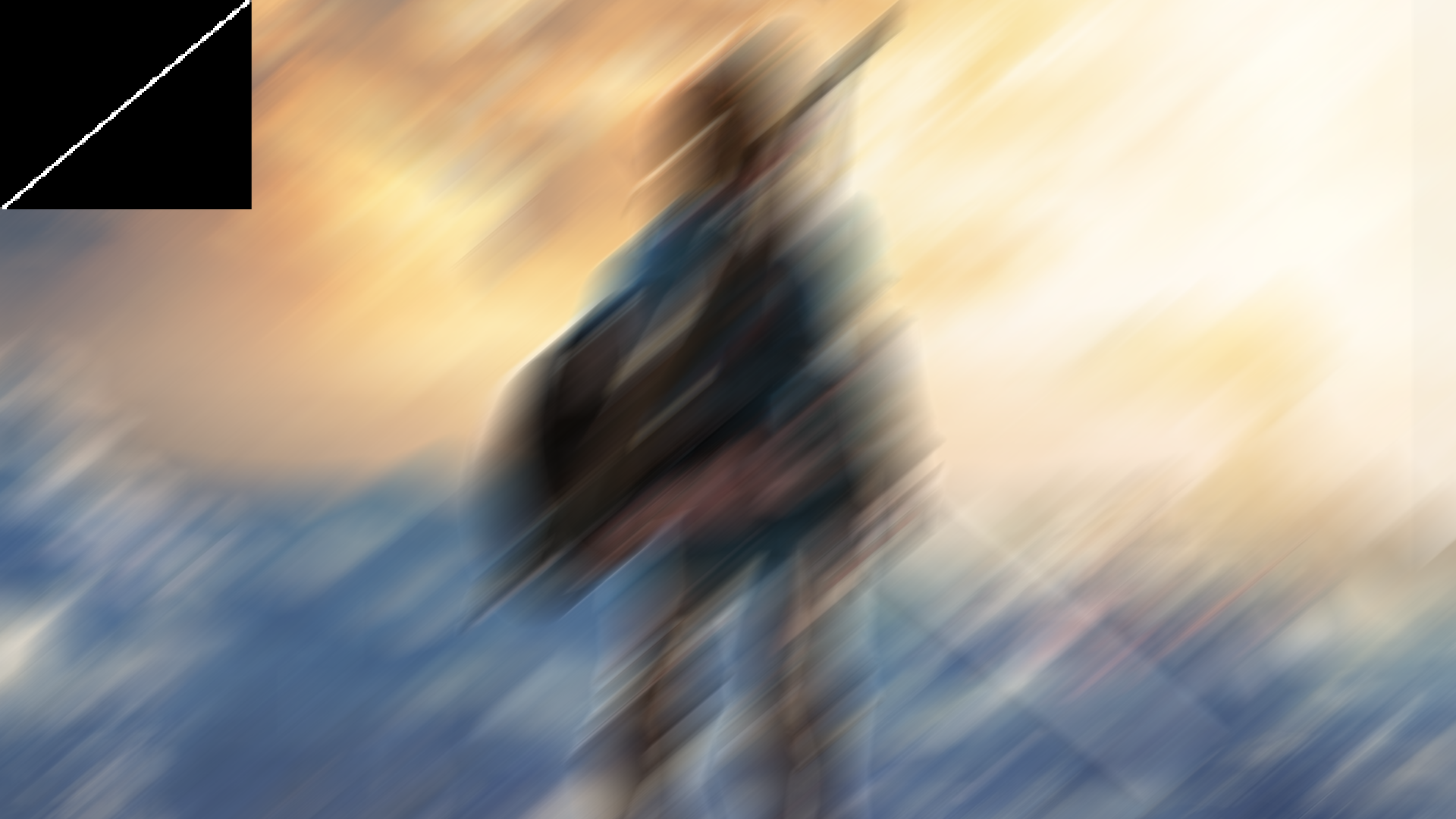}&\includegraphics[width=0.24\linewidth]{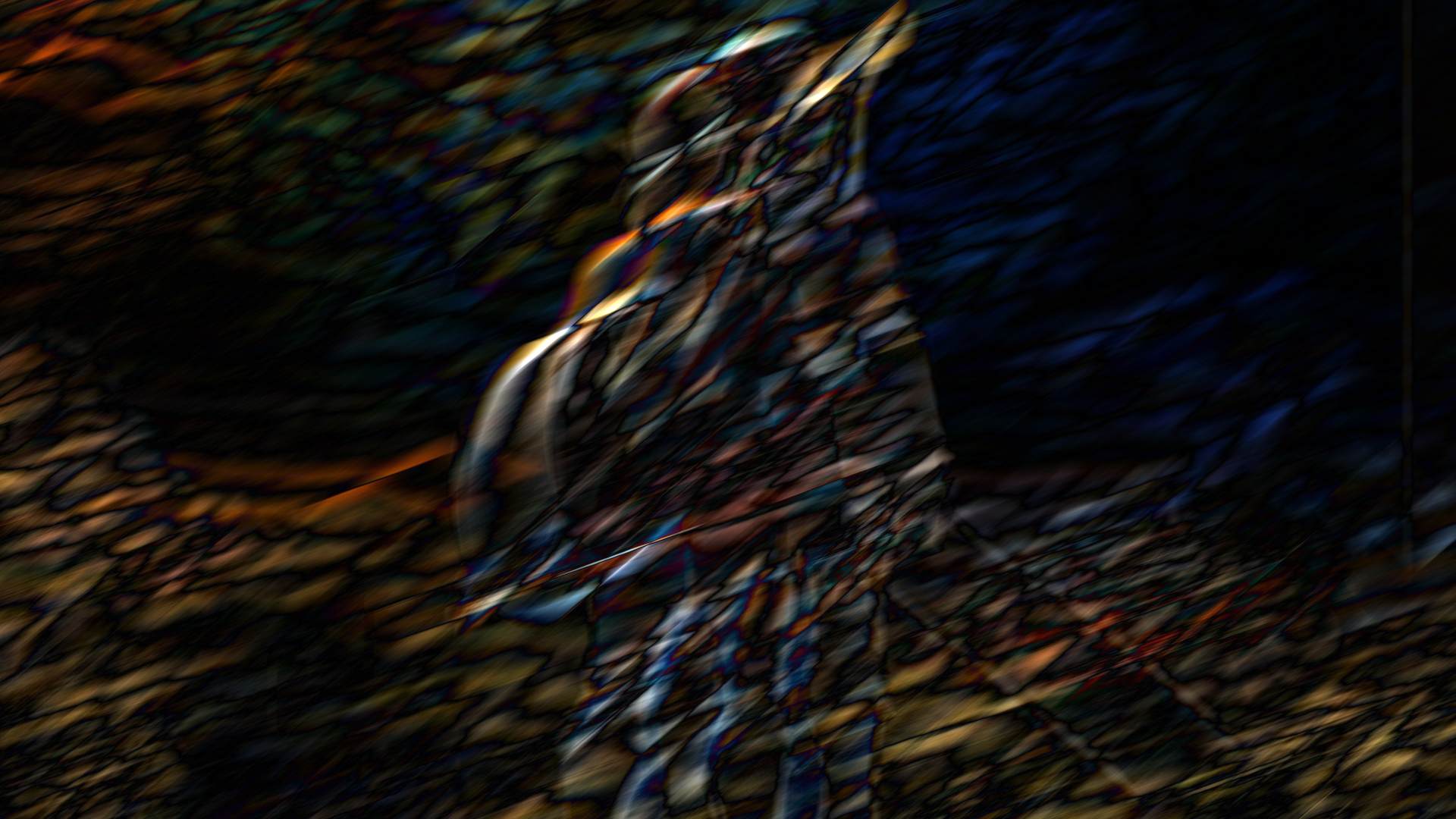}\\
			(a)&(b)&(c)&(d)
		\end{tabular}
		\caption{\textbf{Visualization of the residual induced by the kernel error.} (a) Clear image. (b) Blurry image with the true kernel (motion blur with a length of 150 pixels and orientation of 20$^\circ$). (c) Blurry image with the inaccurate kernel (motion blur with a length of 150 pixels and orientation of 40$^\circ$). (d) The residual induced by kernel error which is the difference between (b) and (c).}
		\label{fig:2}
	\end{figure*}
	\section{Proposed Method}
	In this paper, we propose a dataset-free deep prior for the residual expressed by a customized untrained deep neural network, which allows us to flexibly adapt to different blurs and images in real scenarios. By organically integrating the respective strengths of deep priors and hand-crafted priors, we propose an unsupervised semi-blind deblurring model which recovers the clear image from the blurry image and blur kernel.
	\subsection{Problem Formulation}
	We first take a look at the formulation of the semi-blind deblurring problem.
	Considering the kernel error and possible artifacts, we formulate the degradation process as
	\begin{align}\label{2}
		\boldsymbol{y}&=(\boldsymbol{\widehat k} + \Delta \boldsymbol{k})\otimes \boldsymbol{x}+\bm h+\boldsymbol{n}=\boldsymbol{\widehat k}\otimes \boldsymbol{x}+ \bm r+\bm h+\boldsymbol{n},
	\end{align}\
	where $\boldsymbol{y}$ and $\boldsymbol{x}$ denote the blurry image and clear image respectively, $\boldsymbol{\widehat k}$ and $\Delta \boldsymbol{k}$ represent the inaccurate blur kernel and kernel error, $\boldsymbol{n}$ represents the AWGN, $\otimes$ is the convolution operator, $\boldsymbol{r}=\Delta \boldsymbol{k} \otimes \boldsymbol{x}$ is the residual induced by the kernel error, and $\bm h$ represents the artifacts.

	\subsection{Mathematical Model}
	Note that deriving $\boldsymbol{x}$, {$\boldsymbol{h}$}, and $\boldsymbol{r}$ from the blurry image $\boldsymbol{y}$ is {a typical ill-posed problem} which admits {infinite} solutions. To confine the solution space, prior information of $\boldsymbol{x}$, $\boldsymbol{h}$, and $\boldsymbol{r}$ is required. In this work, we utilize the proposed DRP incorporated with the sparse prior for the residual $\bm r$, deep image prior (DIP) \cite{imaulyanov2018deep} incorporated with total variation for the clear image $\bm x$, and the sparse prior in the discrete cosine transform (DCT) domain for the artifact $\bm h$. The motivations are as follows.
	
	\noindent\textbf{Deep Residual Prior and Sparse Prior in the Spatial Domain for $\bm r$.} 
	We utilize the DRP guided by the sparse prior in the spatial domain to model the residual $\bm r$.
	Since the intrinsic structures of residual are complex and diverse, it is hard to model the residual accurately by hand-crafted priors {or} data-driven priors.  The dataset-free prior expressed by an untrained neural network becomes a natural choice to flexibly and robustly model the complex  residuals. To this end, we propose the dataset-free DRP expressed by a tailored untrained  network---the customized U-Net to capture the residual $\bm r$.   On the other hand, in the spatial domain the residual is generally sparse; see Figure \ref{fig:2}.  Based on this observation, the sparse prior in the spatial domain is utilized as the domain knowledge to guide DRP. These two priors are organically combined to better characterize the complex residual $\bm r$.
	
	\noindent\textbf{Deep Image Prior and Total Variation Prior for $\bm x$.}
	We utilize DIP guided by total variation prior to model the clear image $\bm x$. Since DIP \cite{imaulyanov2018deep,ren2020neural} exploits the statistics of clear image by the structure of an untrained convolutional neural network, it is suitable to model different blurs and images in a zero-shot scenario. On the other hand, to enforce the local smoothness of clear image, we use total variation prior to guide DIP. These two priors are organically combined to better model the clear image $\bm x$. The effectiveness of this combination is verified in the work \cite{ren2020neural}.

	\noindent\textbf{Sparse Prior in the DCT Domain for $\bm h$.}
	We utilize the sparse prior in the DCT domain for modeling the artifacts $\bm h$. It is worth noting that the artifacts {$\boldsymbol{h}$} caused by the kernel error $\Delta \boldsymbol{k}$ generally have strong periodicity around the sharp edges. This indicates that its DCT coefficient $\bm v=\mathcal C\bm h$ ($\mathcal C$ is the DCT operator) is sparse \cite{ji2011robust}, and it is reasonable to consider sparse prior for artifacts $\boldsymbol{h}$ in the DCT domain.
	
	\begin{figure*}[t]
		\centering
		\includegraphics[scale=0.95]{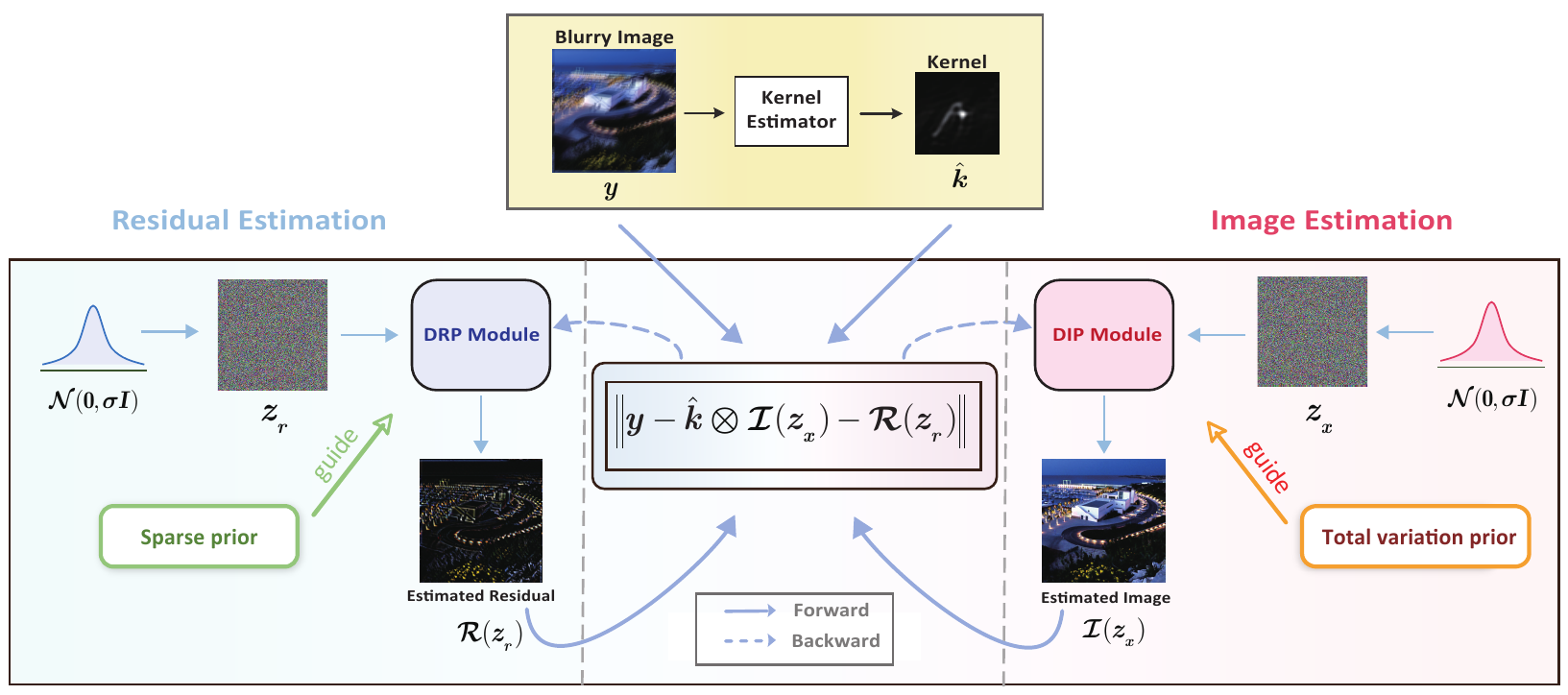}
		\caption{\textbf{The diagram of the proposed model.} The DIP module estimates the image $\mathcal I_{\bm\theta}(\bm{z_x})$ from random noise input $\bm{z_x}$ and the DRP  module estimates the residual $\mathcal R_{\bm\theta}(\bm{z_r})$ from random noise input $\bm {z_r}$.}
		\label{fig3}
	\end{figure*} 
	
	Following the above arguments,  we propose the following unsupervised semi-blind deblurring model (see Figure \ref{fig3} for the diagram). The corresponding optimization problem can be formulated as follows:
	\begin{align}
		\mathop{{\min}}_{\bm{\theta},\bm{\zeta},\bm v}&\left\|\bm y-\boldsymbol{\widehat k}\otimes\mathcal I_{\bm\theta}(\bm{z_x})-\mathcal R_{\bm\zeta}(\bm{z_r})-\mathcal C^T\bm v\right\|_F^2	\nonumber\\+&\lambda_1\|\mathcal I_{\bm\theta}(\bm{z_x})\|_{\text{TV}}+\lambda_2\|\mathcal R_{\bm\zeta}(\bm{z_r})\|_1+\lambda_3\|\bm v\|_1,
	\end{align}
	where $\boldsymbol{y}\in\mathbb R^{n_1\times n_2}$ and $\boldsymbol{x}\in\mathbb{R}^{n_1\times n_2}$ denote the blurry image and clear image, respectively; $\boldsymbol{\widehat k}$  represents the inaccurate blur kernel; $\mathcal I_{\bm\theta}(\bm{z_x})\in\mathbb{R}^{n_1\times n_2} $ and ${\cal R}_{\bm \zeta}(\bm z_{\bm r})\in\mathbb{R}^{n_1\times n_2}$ are the  estimates of the clear image $\bm x$ and the residual $\bm r$, which are generated from untrained neural networks $\mathcal I_{\bm\theta}(\cdot)$ and $\mathcal R_{\bm\zeta}(\cdot)$, respectively; $\bm \theta$ and $\bm \zeta$ collect the corresponding network parameters; $\bm z_x\sim \mathcal N(0,\sigma\bm I)$ and $\bm z_r\sim \mathcal N(0,\sigma\bm I)$  are the random inputs of the neural networks. $\|\cdot\|_{\text{TV}}$ is the total variation regularizer that can be denoted by $\|\nabla_{x}(\cdot)\|_{1}+\|\nabla_{y}(\cdot)\|_{1}$, where  $\nabla_x$ and $\nabla_y$  are the difference operators. Given the inaccurate kernel and the blurry image, we can infer the clear image, residual, and artifacts in model (3) in an unsupervised manner.
	
	In this model, the indispensable hand-crafted priors and deep priors are organicaly intergrated in the our model and work together to deliver promising results.
	
	\subsection{Alternating Minimization Algorithm}
	To tackle the formulated model, we suggest an alternating minimization algorithm. Since the variables are coupled, we suggest decomposing the minimization problem into two easier subproblems by the following alternating minimization scheme:
	\begin{align}
		&\{\bm \theta^{i+1},\bm \zeta^{i+1}\} \in\mathop{\arg\min}_{\bm \theta,\bm \zeta}~\mathcal L\left(\bm \theta,\bm \zeta, {\bm  v }^i \right),\\
		&\bm  v ^{i+1} =\mathop{\arg\min}_{ \bm  v}~\mathcal L\left(\bm \theta^{i+1},\bm  \zeta^{i+1}, \bm v \right), 
		\label{eq:Yupdate}
	\end{align}
	where the superscript ``$i$'' is the iteration number and $\mathcal L(\bm\theta, \bm\zeta, \bm v)$ is the objective function in Eqn. (3). The detailed solutions for these two subproblems are as follows.
	
	\textbf{1) \{$\bm\theta,\bm\zeta$\}-subproblem:} The $\{\bm\theta,\bm\zeta\}$-subproblem (5) can be solved by using gradient-based ADAM algorithm \cite{Adam}. The gradients w.r.t. $\bm \theta$ and $\bm \zeta$ can be computed by the standard back-propagation algorithm \cite{BP}.  Here we jointly update the weights $\bm\theta$ and $\bm \zeta$  in an iteration. 
	
	\textbf{2) $\bm v$-subproblem:}	
	Let 
	\begin{align}
		\mathcal D(\bm\theta,\bm\zeta,\bm v)=\left\|\bm y-\boldsymbol{\widehat k}\otimes\mathcal I_{\bm\theta}(\bm{z_x})-\mathcal R_{\bm\zeta}(\bm{z_r})-\mathcal C^T\bm v\right\|_F^2, \nonumber
	\end{align}
	then the $\bm v$-subproblem (5) is 
	\begin{align}
		\bm v^{i+1}=\mathop{\arg\min}_{\bm v} ~\lambda_3\|\bm v\|_1+\mathcal D(\bm\theta^{i+1},\bm\zeta^{i+1},\bm v)\nonumber,
	\end{align}
	which can be exactly solved by using proximal gradient descent: 
	\begin{align}
		\bm v^{i+1}=\mathcal S_{\lambda_3/L}\left(\bm v^i-\frac{1}{L}\nabla_{\bm v}\mathcal D(\bm\theta^{i+1},\bm\zeta^{i+1},\bm v^i)\right),
	\end{align} 
	where $L>0$ is a constant, and {$\mathcal S_\delta$} is the soft-thresholding operator defined by
	$$\mathcal{S}_{\delta}(\bm a)_{ij}=\max\left(\left|a_{ij}\right|-\delta, 0\right)\operatorname{sgn}\left(a_{ij}\right).$$
	where $a_{ij}$ is the $(i,j)$-th entry of matrix $\bm a$. The overall minimization algorithm is summarized in Algorithm 1.

	\begin{minipage}{8cm}
		\begin{algorithm}[H]
			\caption{Alternating minimization algorithm}  
			\textbf{Input}:  Blurry image $\bm y$,  inaccurate kernel $\bm{\widehat k}$, the parameters ${\lambda_s~(s=1,2,3)}$, and iteration number $T$.  \\
			\textbf{Initailization}: Random input $\bm z_x\sim \mathcal N(0,\sigma\bm I)$ and $\bm z_r\sim \mathcal N(0,\sigma\bm I)$.
			
			\begin{algorithmic}[1]
				\FOR{$i=0$ to $T$}
				\STATE $\widehat{\bm{x}}^{i+1}=\mathcal I_{\bm\theta^{i}}(\bm{z_x}), 
				\widehat{\bm{r}}^{i+1}=\mathcal R_{\bm\zeta^{i}}(\bm{z_r})$;
				\STATE compute the gradients w.r.t $\bm\theta$ and $\bm\zeta$;
				\STATE jointly update $\bm\theta^{i+1}, \bm\zeta^{i+1}$ using ADAM;
				\STATE update $\bm v^{i+1}$ using (6);			
				\ENDFOR
			\end{algorithmic} 
			\textbf{Output}: the restored image $\widehat{\bm{x}}$ and the residual $\widehat{\bm{r}}.$
		\end{algorithm} 
	\end{minipage}
	
	\section{Experiments}
	Extensive experiments on simulated and real blurry images are reported to substantially demonstrate the effectiveness of the proposed method, especially the robustness to different types of kernel error.
	\subsection{Experimental Setup}
	
	\begin{itemize}	
		
		\item \textbf{Evaluation Metrics.}
		Following prior works of deblurring, we use the peak signal-to-noise ratio (PSNR) and structural similarity index (SSIM) to evaluate the restoration quality of the images. To evaluate the quality of residual estimation, we use the mean square error (MSE), which measures the average difference of pixels in the entire true residual and estimated residual. A lower MSE value means there is less difference between the true residual and estimated residual.
		
		\item \textbf{Model Hyperparameters Setting.} The model hyperparameters of the proposed method are $\{\lambda_i\}_{i=1}^{3}$, which are tuned to obtain best PSNR value.  In our experiments, $\lambda_1,~\lambda_2,~\lambda_3$ are empirically determined to be $5\times10^{-2},5\times10^{-5},5\times10^{-7}$, respectively. We also conduct the sensitivity analysis of different model hyperparameters in the \textbf{supplementary materials}.
		\item \textbf{Algorithm Hyperparameters Setting.} In all experiments, we set the total iteration number of the proposed algorithm to be 1500.  The default learning rates of networks $\mathcal I_{\bm\theta}$ and $\mathcal R_{\bm\zeta}$ are respectively $9\times10^{-3}$ and $5\times10^{-4}$. 
		\item \textbf{Platform.} In this work, all experiments are conducted on the PyTorch 1.10.1 and MATLAB 2017b platform with an i5-12400f CPU, RTX 3060 GPU, and 16GB RAM.
	\end{itemize}
	\subsection{Ablation Study}
	The ablation studies primarily focus on two aspects: {\it a}) exploring the influence of different combinations of deep priors and hand-crafted priors; {\it b}) exploring the superiority of the customized U-Net compared to other classic architectures in capturing the residual $\bm r$ induced by kernel error. 
	\subsubsection{The Influence of Different Combinations of Deep Priors and Hand-crafted Priors} 
	\begin{table}[!h]
		\centering
		\caption{The average metric values of the restored results by reconciling different hand-crafted priors with deep priors.}
		{
			\renewcommand{\arraystretch}{0.9}
			\begin{tabular}{c|cc}
				\toprule
				Method&PSNR&SSIM\\\midrule
				w/o sparse prior for $\bm r$&26.77&0.853\\
				w/o sparse prior for $\bm v$&27.25&0.845\\
				w/o total variation for $\bm x$&27.98&0.871\\
				w/o DIP for $\bm x$ &26.54&0.867\\
				w/o DRP for $\bm r$&25.69&0.842\\
				Ours&\textbf{28.64}&\textbf{0.892}\\\bottomrule
			\end{tabular}
		}
	\end{table}
	To investigate the role of different priors in our model, we perform experiments by reconciling deep priors and hand-crafted priors. The test images contain 24 blurry images that are blurred with 8 sharp images  and  3 typical kernels; see \textbf{supplementary materials}.
	Table 1 lists the average metric values of the restored results by reconciling different hand-crafted priors with deep priors. We can observe from  the table that the deep priors and hand-crafted priors are indispensable, working together to deliver promising results. 
	
	\subsubsection{The Influence of Different Network Architectures for Deep Residual Prior} 
	\begin{figure}[t]
		\centering
		\includegraphics[scale=0.41]{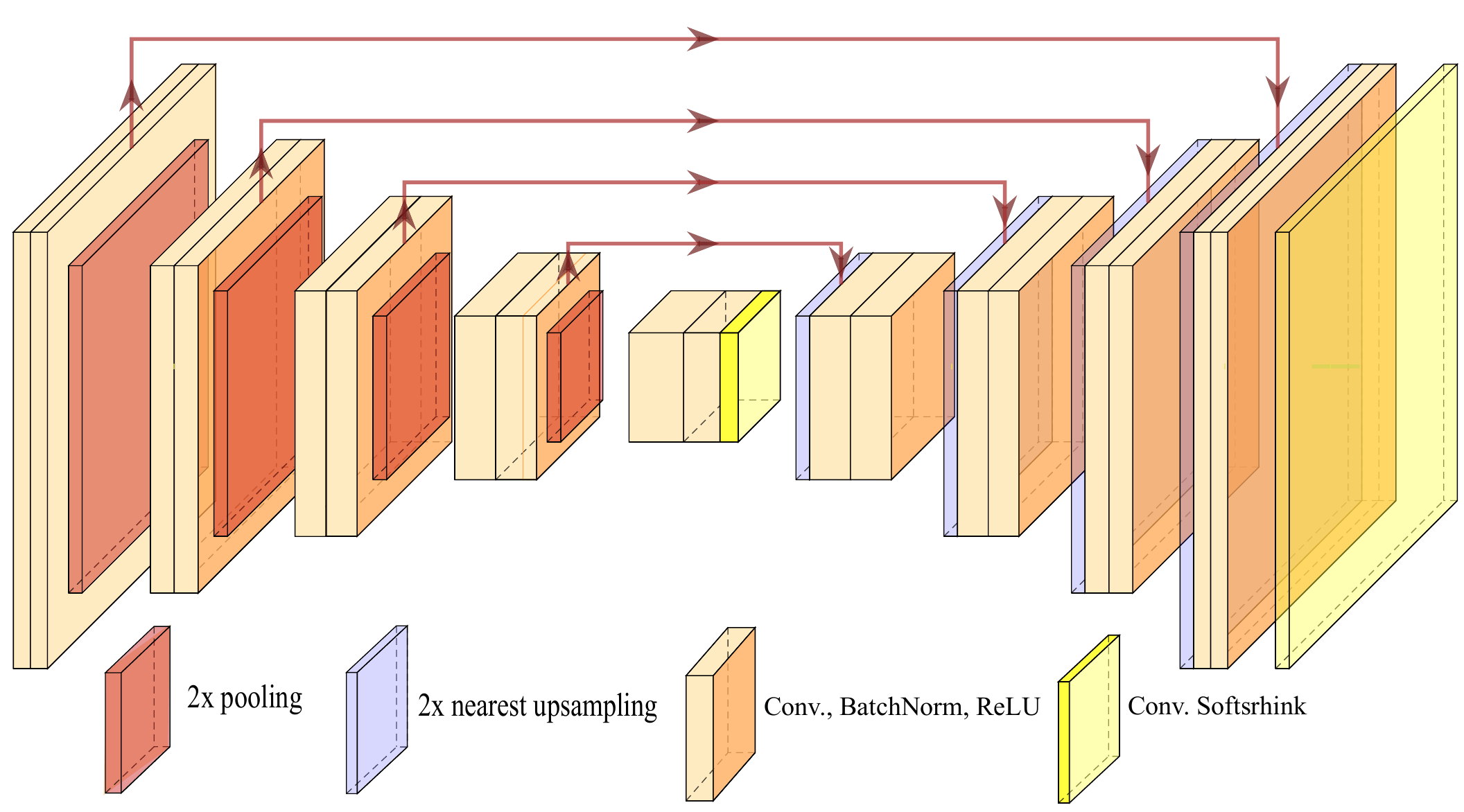}
		\caption{The customized U-Net used in our method, which maps the random noise input $\bm{z_r}$ to the residual $\bm r$.} 
		\label{struc}
	\end{figure}
	\begin{figure}[h]
		\centering
		\setlength\tabcolsep{1pt}
		\begin{tabular}{cccc}
			\includegraphics[width=0.24\linewidth]{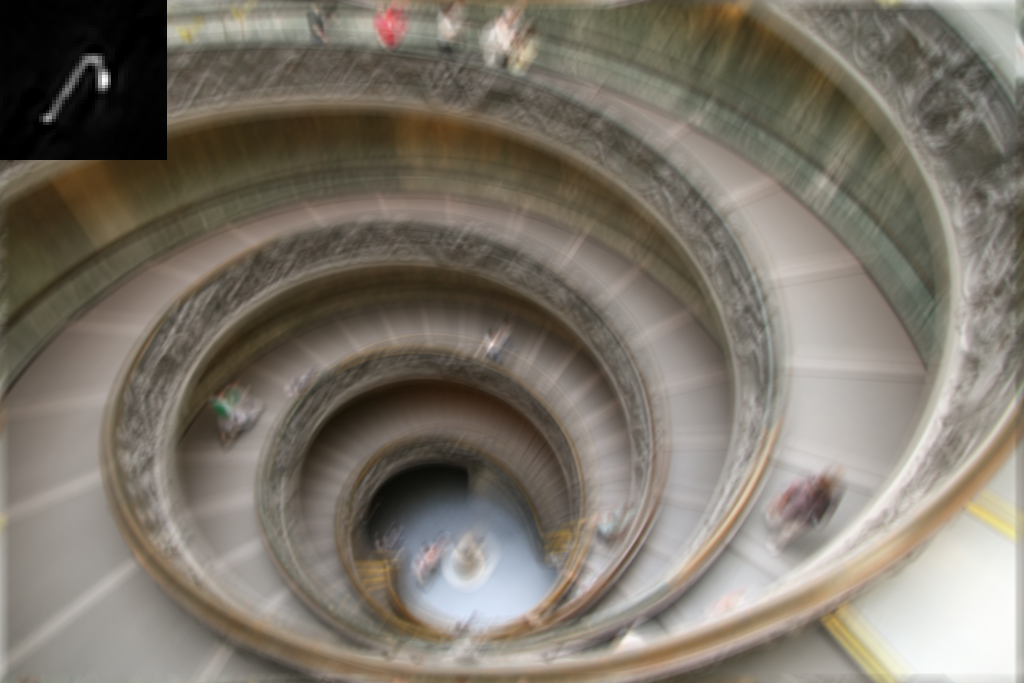}&\includegraphics[width=0.24\linewidth]{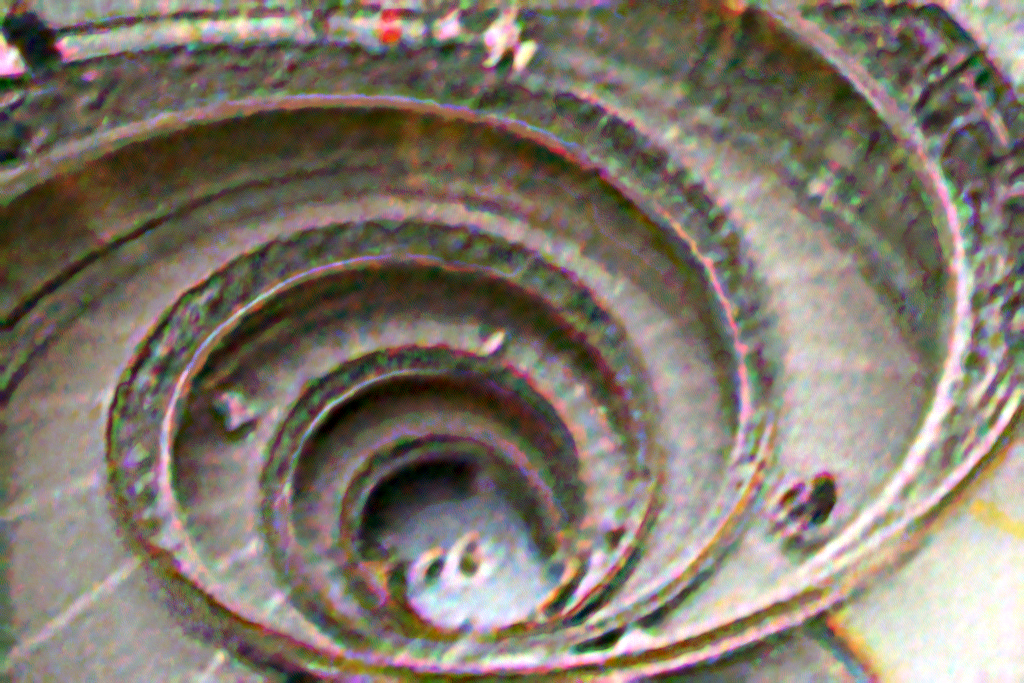}&\includegraphics[width=0.24\linewidth]{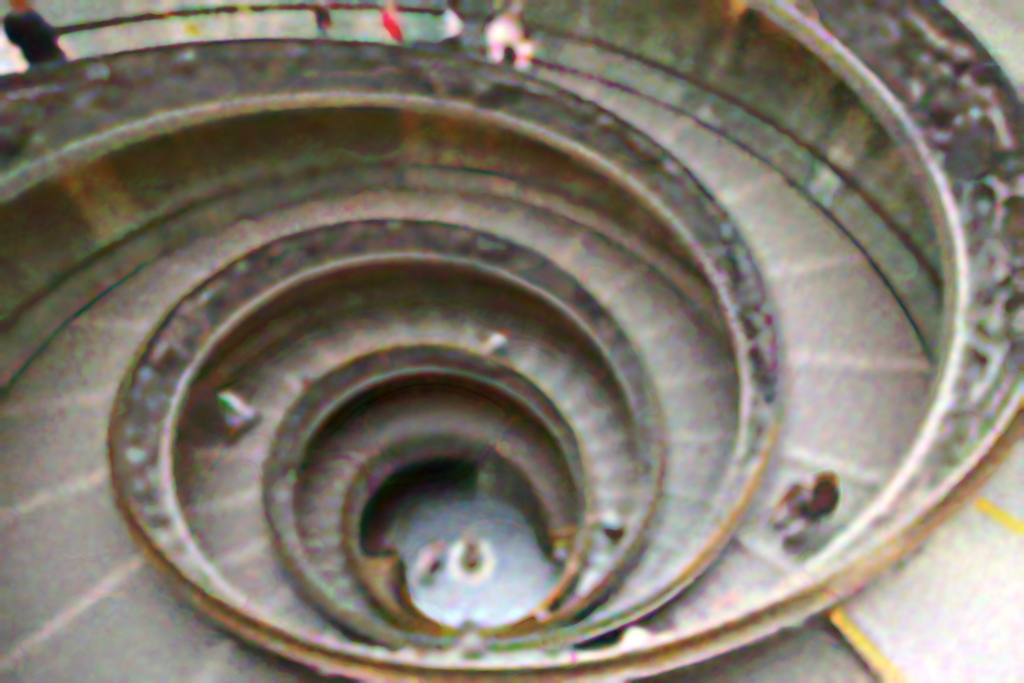}&\includegraphics[width=0.24\linewidth]{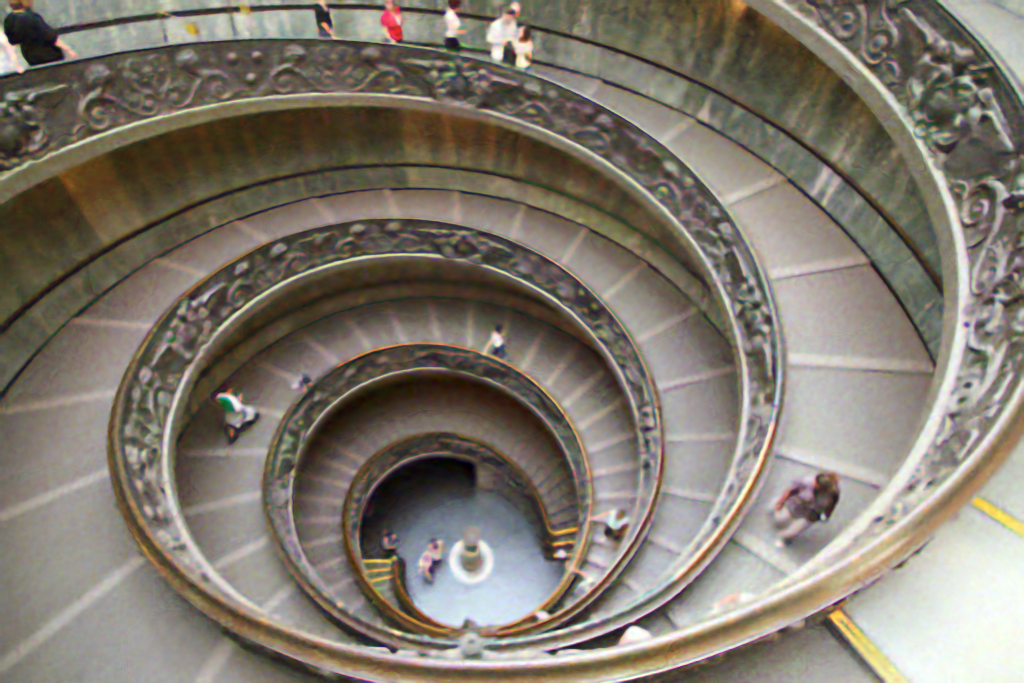}\\
			Blurry&PSNR 18.32&PSNR 21.82&PSNR \textbf {24.02}\\
			\includegraphics[width=0.24\linewidth]{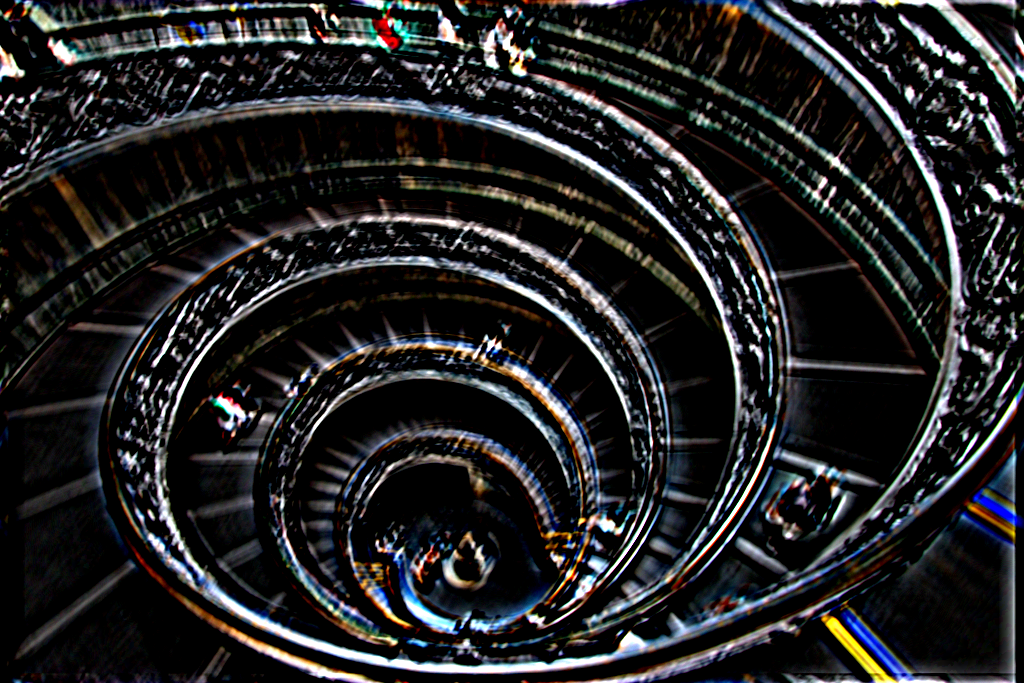}&\includegraphics[width=0.24\linewidth]{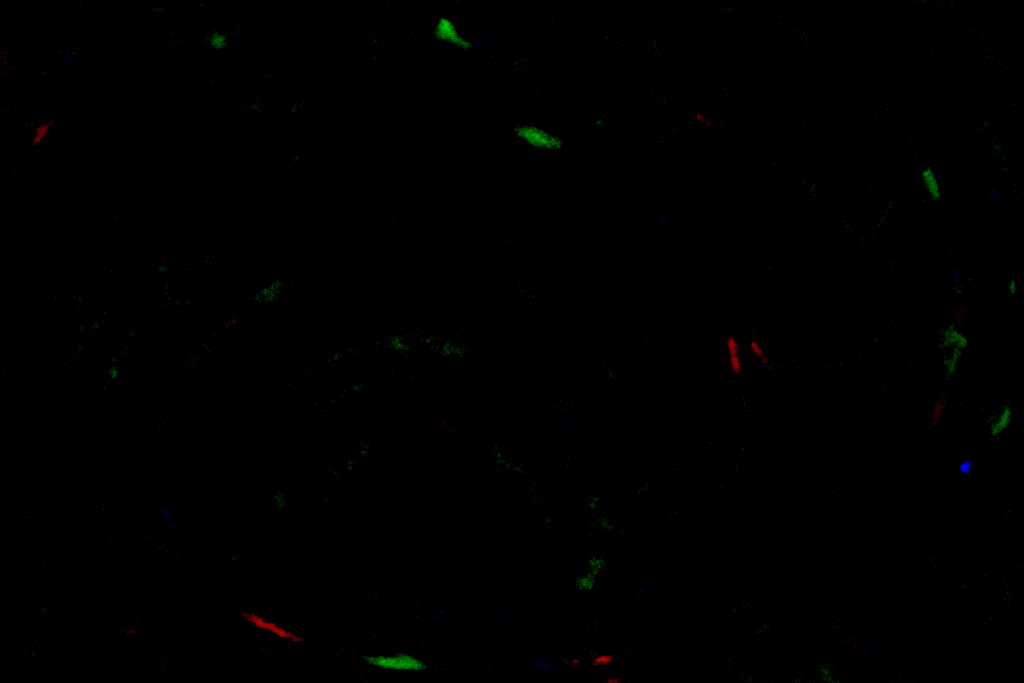}&\includegraphics[width=0.24\linewidth]{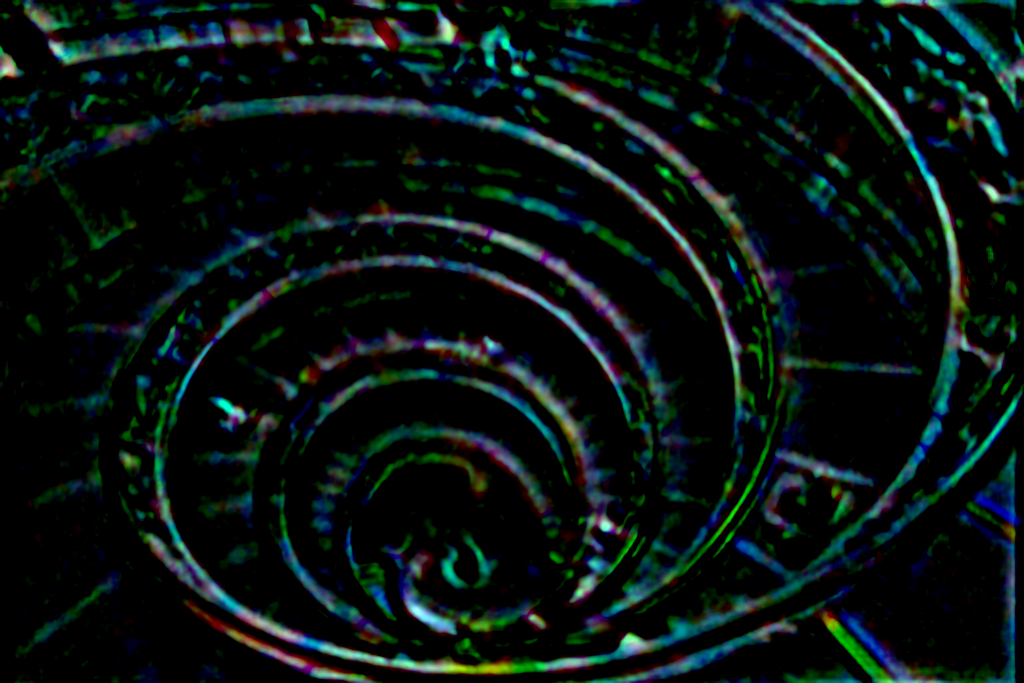}&\includegraphics[width=0.24\linewidth]{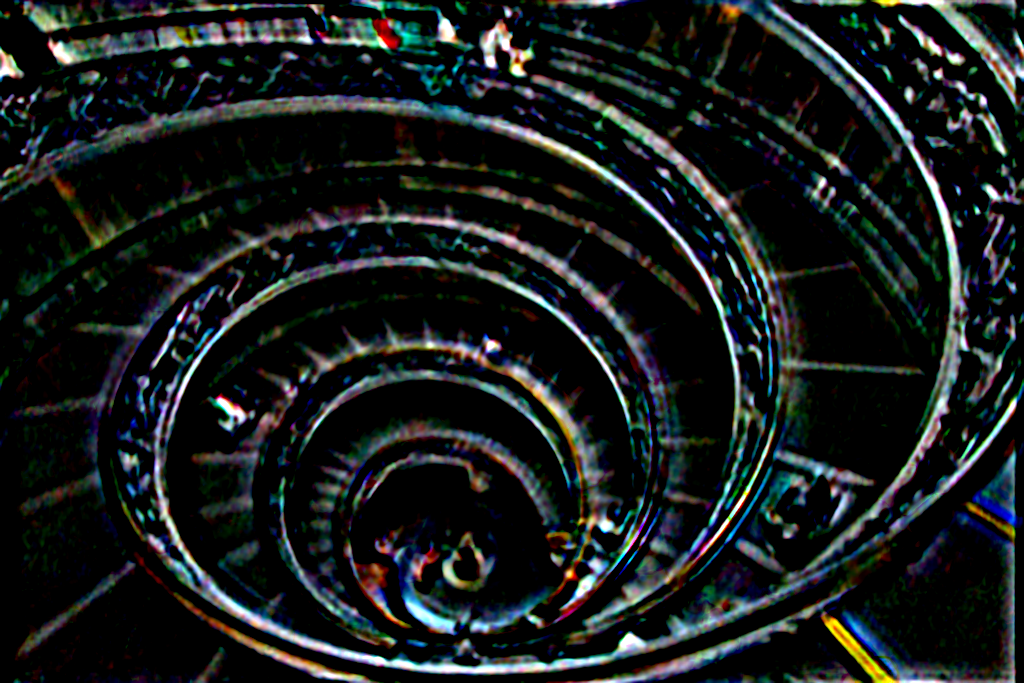}\\
			MSE 0&MSE 0.0274&MSE 0.0126&MSE \textbf{0.0046}\\
			True Res.&FCN&U-Net&Ours
		\end{tabular}
		\caption{\textbf{The influence of different network architectures for DRP.} Row 1: the deblurred results of the image in the dataset of \cite{lai2016comparative}. Row 2: the estimated residuals. The input kernel for deblurring is estimated by the method in \cite{perrone2014total}. The closer estimated residual is to the true residual, the better it is.}\label{Structure}
	\end{figure}	 
	We discuss the influence of different network architectures for DRP. Three network architectures are used for comparison: (i) fully connected network (FCN), (ii) U-Net, and (iii) the customized U-Net (see Figure \ref{struc}), in which the soft-shrinkage is used to replace the sigmoid.	 Figure \ref{Structure} displays the deblurred results and estimated residuals by using different network architectures. We can observe from the results that the customized U-Net is more favorable to capture the residual as compared to FCN and  U-Net. The vanishing gradient issue is encountered for the sigmoid function when training  neural networks  for the  residual with subtle values. Thus, the customized  U-Net, which replaces the sigmoid function with the soft-shrinkage function, can better capture the residual as the soft-shrinkage can ameliorate this issue as compared to the sigmoid function. Consequently, we can observe from Figure \ref{Structure} that a better estimation of the residual boosts the quality of image restoration.
	
	\subsection{Performance on Benchmark Datasets}	
	We comprehensively compare our method with state-of-the-art methods on simulated data and real data. The simulated data comes from the datasets of  Levin \etal\cite{levin2009understanding} and Lai \etal\cite{lai2016comparative}, in which the Gaussian noise level  is set to be 1\%. The real data comes from the dataset of Lai \etal\cite{lai2016comparative}. The dataset of Levin \etal\cite{levin2009understanding} contains 32 blurry images that are blurred with 4 sharp images and 8 blur kernels. The dataset of Lai \etal\cite{lai2016comparative} contains 102 real blurry images and 100 simulated blurry images that are blurred with 25 sharp images and 4 blur kernels. 
	
	\noindent\textbf{Kernel Estimation.} The estimated kernels for the dataset of Levin \etal\cite{levin2009understanding} are obtained by applying five blind deblurring algorithms, including Cho \etal\cite{cho2009fast}, SelfDeblur by Ren \etal~\cite{ren2020neural}, Levin \etal~\cite{levin2009understanding}, Pan \etal~\cite{pan2016blind}, and Sun \etal~\cite{sun2013edge}. The estimated kernels for the dataset of Lai \etal~\cite{lai2016comparative} are obtained by four blind deblurring methods, including Xu \etal~\cite{xu2010two}, Xu \etal~\cite{xu2013unnatural}, Sun \etal~\cite{sun2013edge}, and Perrone \etal~\cite{perrone2014total}. Note that, two groups of different kernel estimation algorithms are chosen for  two datasets respectively to follow the settings in the compared methods.
	\begin{figure}[htbp]
		\setlength\tabcolsep{1pt}
		\renewcommand{\arraystretch}{1}  
		\centering
		\begin{tabular}{ccccc}
			\includegraphics[width=0.195\linewidth]{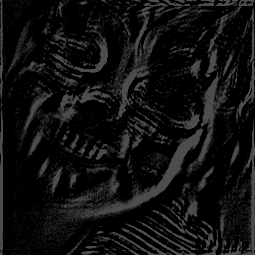}&
			\includegraphics[width=0.195\linewidth]{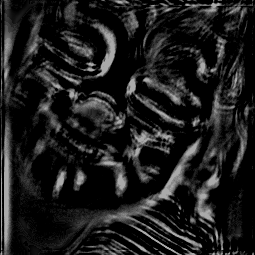}&
			\includegraphics[width=0.195\linewidth]{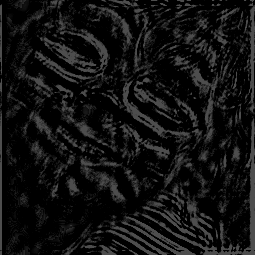}&
			\includegraphics[width=0.195\linewidth]{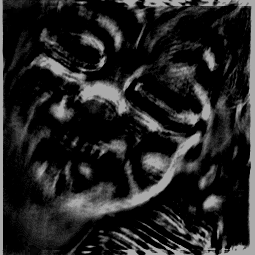}&
			\includegraphics[width=0.195\linewidth]{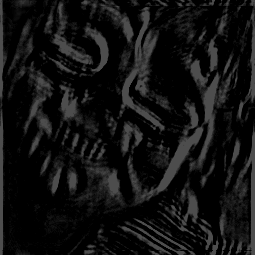}
			\\
			MSE 0&MSE 0.012&MSE 0.003&MSE 0.026&MSE \textbf{0.002} \\
			\includegraphics[width=0.195\linewidth, height=0.195\linewidth]{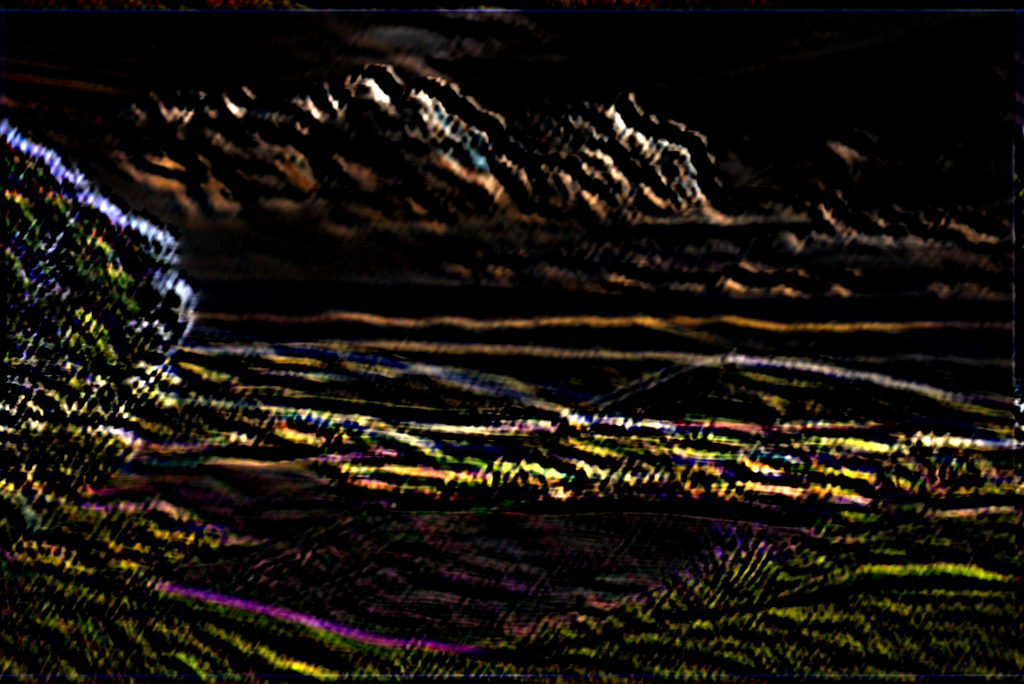}&
			\includegraphics[width=0.195\linewidth,height=0.195\linewidth]{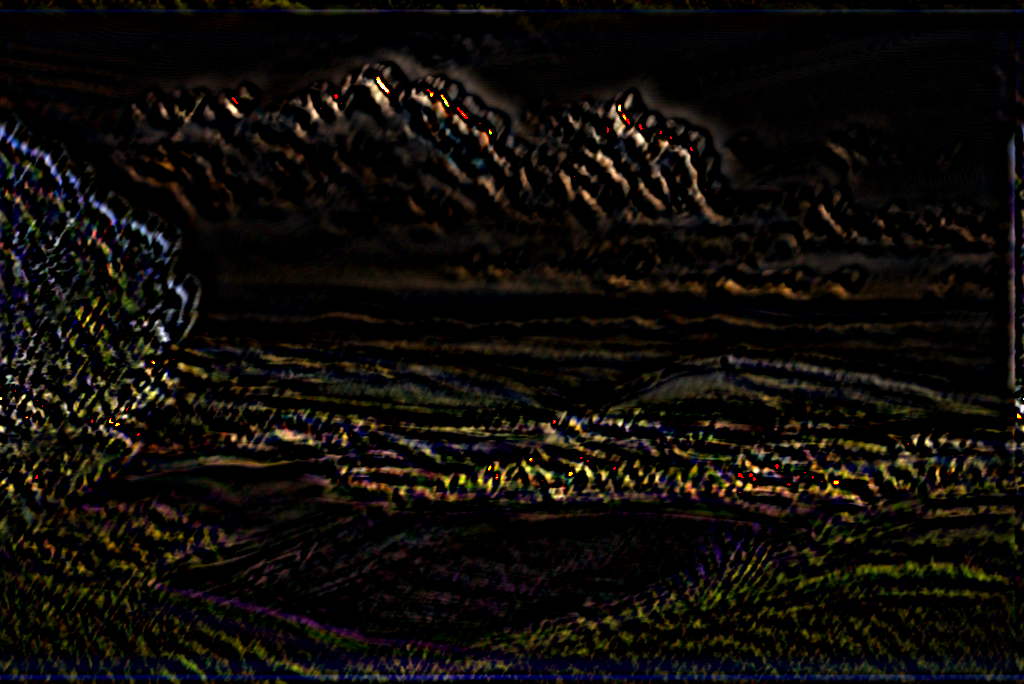}&
			\includegraphics[width=0.195\linewidth,height=0.195\linewidth]{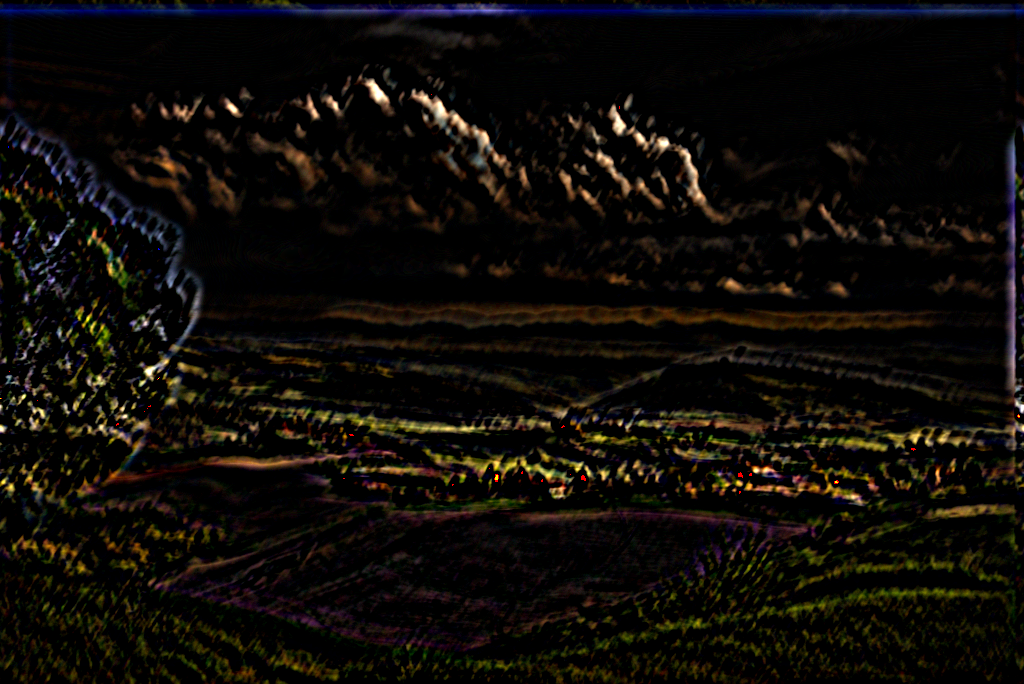}&
			\includegraphics[width=0.195\linewidth,height=0.195\linewidth]{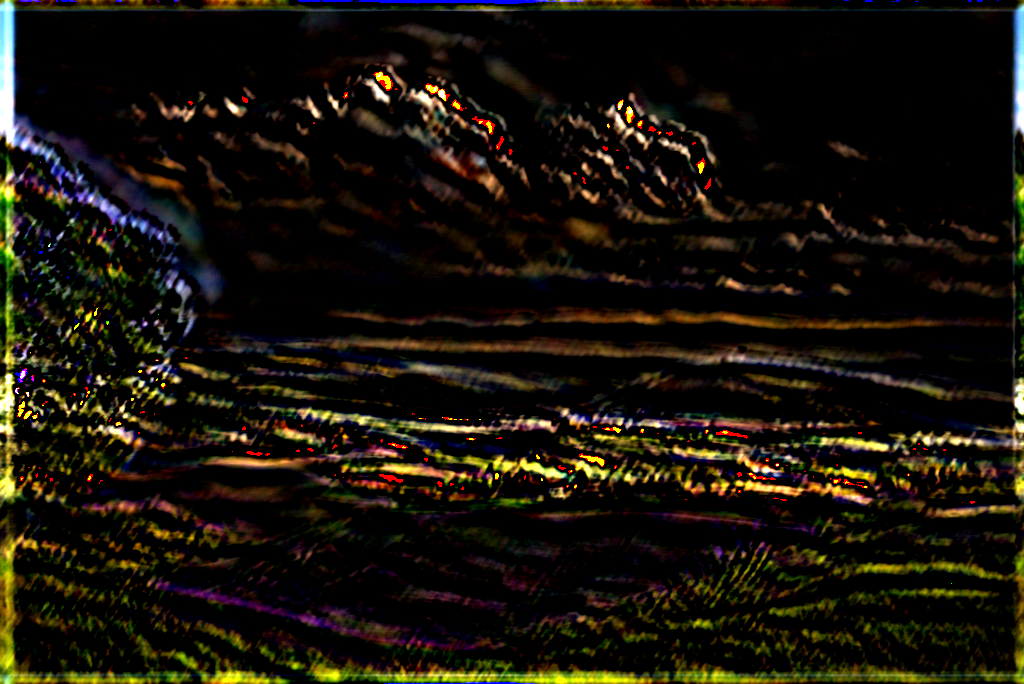}&
			\includegraphics[width=0.195\linewidth,height=0.195\linewidth]{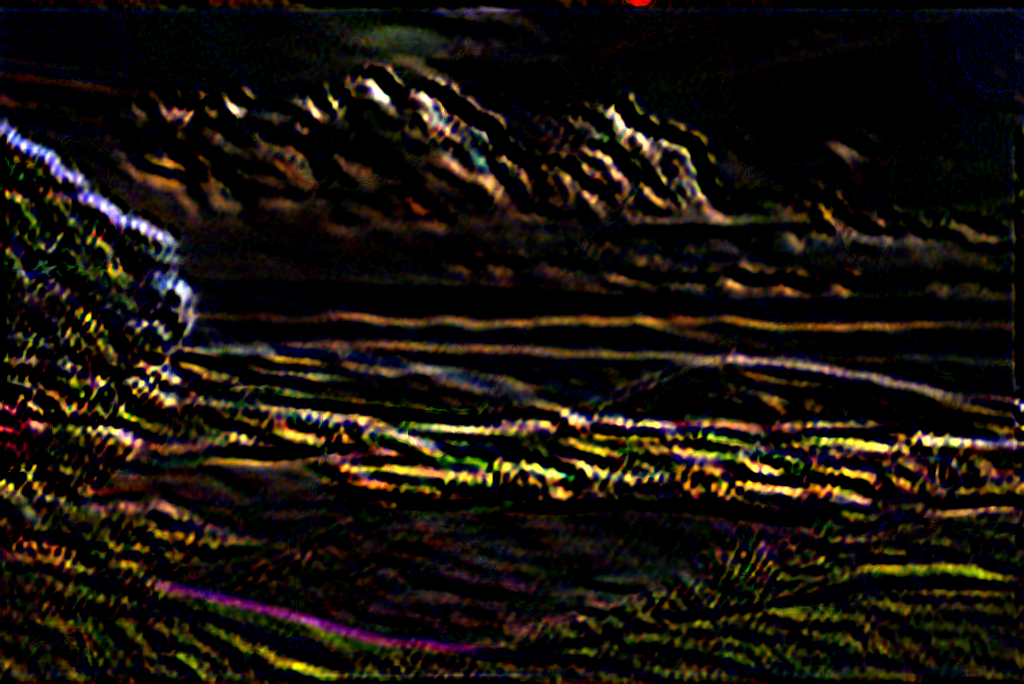}
			\\
			MSE 0&MSE 0.050&MSE 0.047&MSE 0.020&MSE \textbf{0.009}\\
			True Res.&Res. in \cite{ji2011robust}&Res. in \cite{vasu2018non} &Res. in \cite{Fang_2022_CVPR} & Ours
		\end{tabular}
		\caption{The estimated residual by semi-blind methods on the images from datasets of Levin \etal\cite{levin2009understanding} and Lai \etal\cite{lai2016comparative}.  The closer estimated residual is to the true residual, the better it is.}
		\label{reslevin}
	\end{figure} 
	\begin{table*}[htbp]
		\centering
		\caption{Average PSNR/SSIM comparison of deblurred results by different methods on the dataset of Levin \etal\cite{levin2009understanding} and Lai \etal\cite{lai2016comparative} using different estimated kernels.}
		\resizebox{\linewidth}{!}{
			\renewcommand{\arraystretch}{1.2}
			\begin{tabular}{c|c|c|ccc|ccccc}
				\bottomrule[1pt]&&\multicolumn{1}{c|}{BD}&\multicolumn{3}{c|}{NBD}&\multicolumn{4}{c}{SBD}\\\hline
				Datasets&\diagbox{Kernel}{Method}&Ren \etal\cite{ren2020neural}&Krishnan \etal\cite{krishnan2009fast}&Ren \etal\cite{ren2015fast}&Eboli \etal\cite{eboli2020end2end}&Zhao \etal\cite{Zhao2013TotalVS}&Ji \& Wang\cite{ji2011robust}&Vasu \etal\cite{vasu2018non}&Fang \etal\cite{Fang_2022_CVPR}&Ours\\ \hline
				\multirow{6}{*}{Levin \etal\cite{levin2009understanding}}&Cho \etal\cite{cho2009fast}&\multirow{6}{*}{28.52/0.80}& 28.12/0.82& 28.49/0.81&28.54/0.82&28.76/0.82&28.55/0.83 &29.50/0.84&29.86/0.85&\textbf{30.52/0.88}\\
				&Levin \etal\cite{levin2011efficient} &&28.17/0.83&28.55/0.83&28.93/0.84&27.96/0.81&28.79/0.83&29.65/0.85&29.91/0.86&\textbf{30.78/0.89}\\
				&Ren \etal\cite{ren2020neural}&&27.88/0.82&27.59/0.80&27.74/0.82&27.19/0.80&27.39/0.82&29.01/0.85&28.57/0.83&\textbf{29.55/0.86}\\
				&Pan \etal\cite{pan2016blind} &&29.84/0.86&29.99/0.86&30.58/0.87&28.98/0.83&29.21/0.86&31.62/0.87&31.98/0.91&\textbf{32.28/0.92}\\
				&Sun \etal\cite{sun2013edge}&&29.30/0.85&29.91/0.86&30.90/0.87&28.99/0.83&29.10/0.86&30.75/0.88&31.05/0.89&\textbf{31.84/0.91}\\
				\cline{2-11}
				&Average&28.52/0.80&28.66/0.83&28.91/0.83&29.34/0.84&28.38/0.82&28.61/0.84&30.10/0.86&30.27/0.87&\textbf{31.01/0.89}\\
				\hline
				\multirow{5}{*}{Lai \etal\cite{lai2016comparative}}&Xu \etal\cite{xu2010two} &\multirow{4}{*}{17.69/0.64}&17.73/0.65&17.61/0.56&19.53/0.64&19.60/0.64&19.67/0.61&20.21/0.67&20.67/0.68&\textbf{21.07/0.68}\\
				&Xu \etal\cite{xu2013unnatural}& &18.83/0.62&17.81/0.57&19.52/0.67&19.78/0.67&19.34/0.60&19.52/0.64&20.25/0.66&\textbf{20.96/0.68}\\
				&Sun \etal\cite{sun2013edge}&&17.96/0.61&17.99/0.57&20.04/0.69&19.83/0.69&20.10/0.68&19.88/0.67&20.44/0.68&\textbf{21.03/0.70}\\
				&Perrone \etal\cite{perrone2014total} &&17.57/0.63&16.88/0.54&18.89/0.62&19.56/0.65&19.21/0.60&19.12/0.60&19.87/0.62&\textbf{20.31/0.64}\\
				\cline{2-11}
				&Average&17.69/0.64&18.02/0.63&17.57/0.56&19.49/0.66&19.69/0.66&19.58/0.63&19.68/0.65&20.31/0.66&\textbf{20.84/0.68}\\
				\toprule[1pt]
			\end{tabular}
		}
	\end{table*}
	
	\noindent\textbf{Compared Methods.} We select eight state-of-the-art methods as the baselines. These methods cover a blind deblurring (BD) method SelfDeblur \cite{ren2020neural} (in which the authors of \cite{ren2020neural} use the Levin \etal's dataset and Lai \etal's dataset without adding extra noise while the noise level  in our setting is 1\%), three non-blind (NBD) deblurring methods (i.e., Fergus \etal\cite{fergus2006removing}, Ren \etal\cite{ren2015fast}, and Eboli \etal\cite{eboli2020end2end}), and four semi-blind deblurring (SBD) methods (i.e., Zhao \etal \cite{Zhao2013TotalVS}, Ji and Wang \cite{ji2011robust}, Vasu \etal \cite{vasu2018non}, Fang \etal \cite{Fang_2022_CVPR}). 
	\begin{figure*}[!t]
		\setlength\tabcolsep{1pt}
		\renewcommand{\arraystretch}{0.5} 
		\centering
		\begin{tabular}{ccccccccc}
			\includegraphics[width=0.105\linewidth]{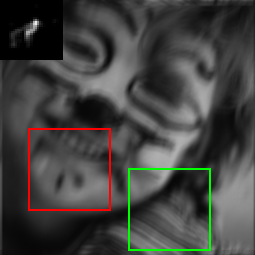}&
			\includegraphics[width=0.105\linewidth]{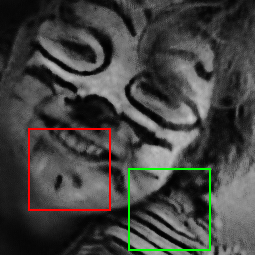}&
			\includegraphics[width=0.105\linewidth]{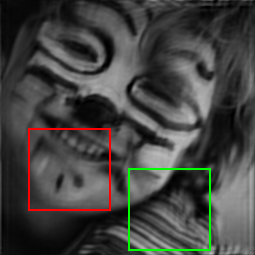}&
			\includegraphics[width=0.105\linewidth]{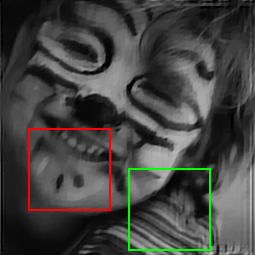}&
			\includegraphics[width=0.105\linewidth]{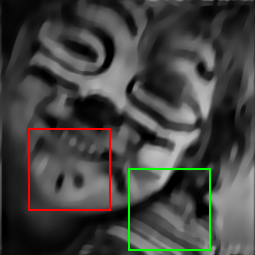}&
			\includegraphics[width=0.105\linewidth]{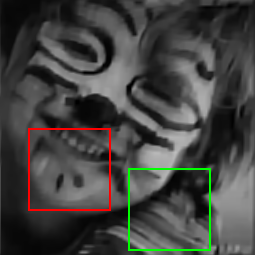}&
			\includegraphics[width=0.105\linewidth]{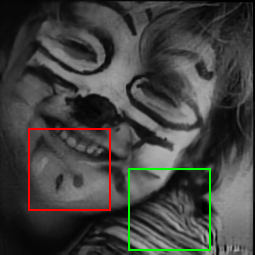}&
			\includegraphics[width=0.105\linewidth]{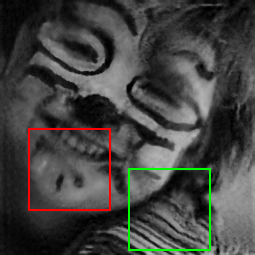}&
			\includegraphics[width=0.105\linewidth]{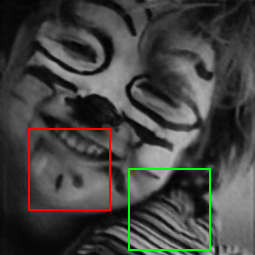}\\
			\includegraphics[width=0.105\linewidth]{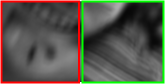}&
			\includegraphics[width=0.105\linewidth]{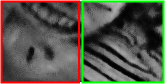}&
			\includegraphics[width=0.105\linewidth]{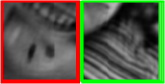}&
			\includegraphics[width=0.105\linewidth]{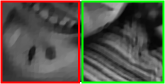}&
			\includegraphics[width=0.105\linewidth]{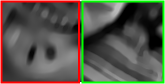}&
			\includegraphics[width=0.105\linewidth]{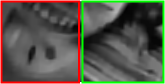}&
			\includegraphics[width=0.105\linewidth]{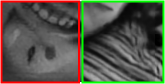}&
			\includegraphics[width=0.105\linewidth]{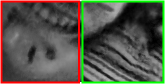}&
			\includegraphics[width=0.105\linewidth]{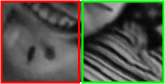}\\
			24.64/0.747&25.96/0.765&29.57/0.878&29.03/0.870&26.21/0.774&28.28/0.856&30.25/0.877&29.00/0.875&\textbf{31.65/0.889}\\
			\includegraphics[width=0.105\linewidth,]{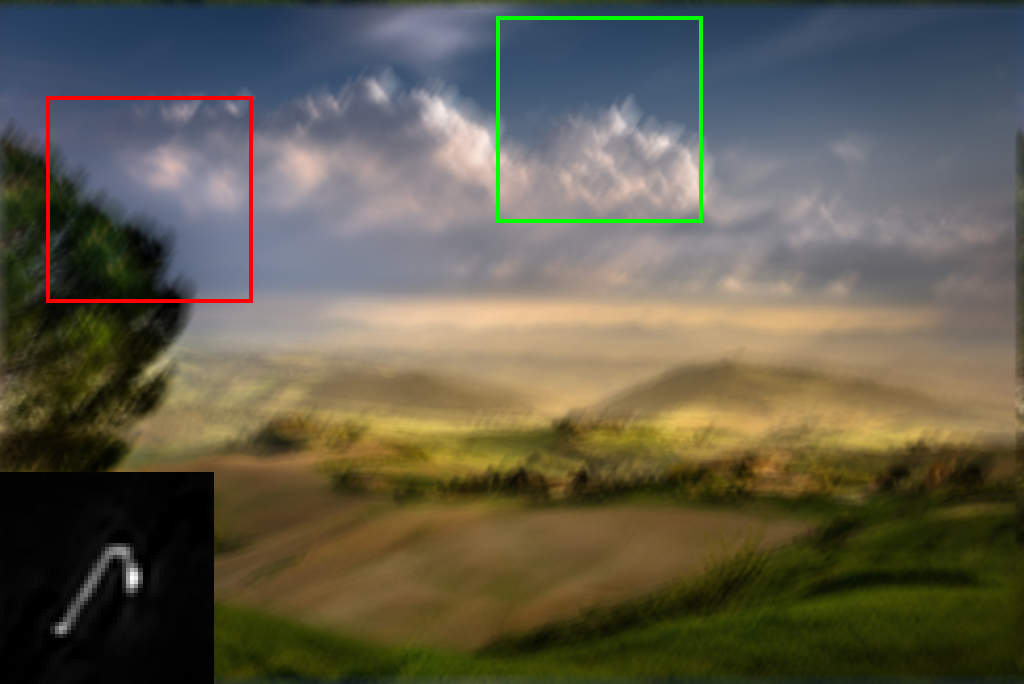}&
			\includegraphics[width=0.105\linewidth,]{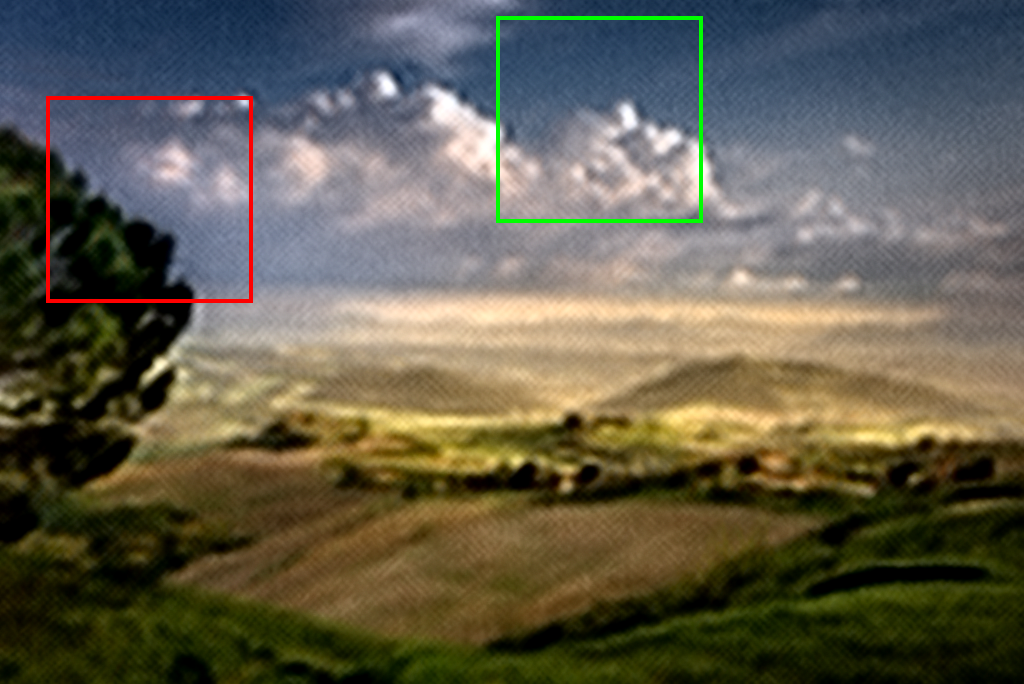}&
			\includegraphics[width=0.105\linewidth,]{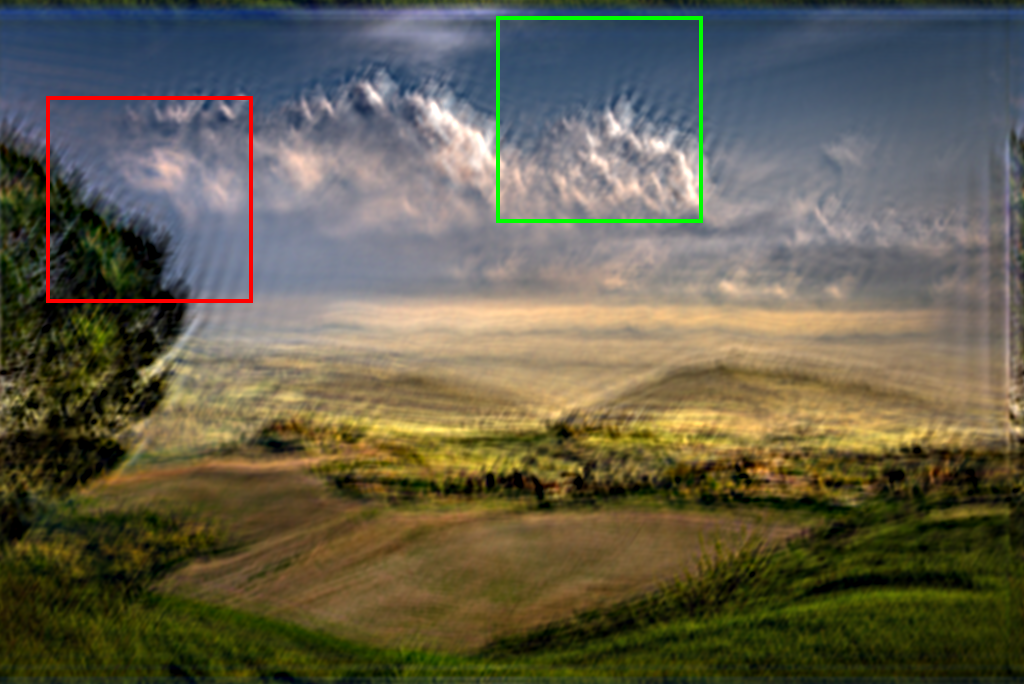}&
			\includegraphics[width=0.105\linewidth,]{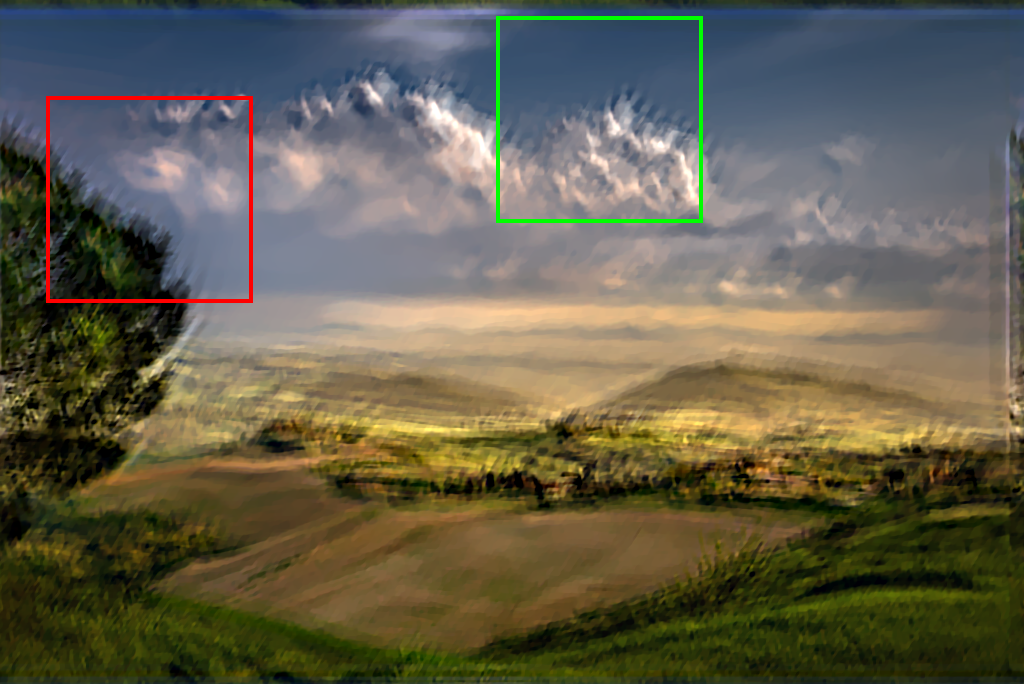}&
			\includegraphics[width=0.105\linewidth,]{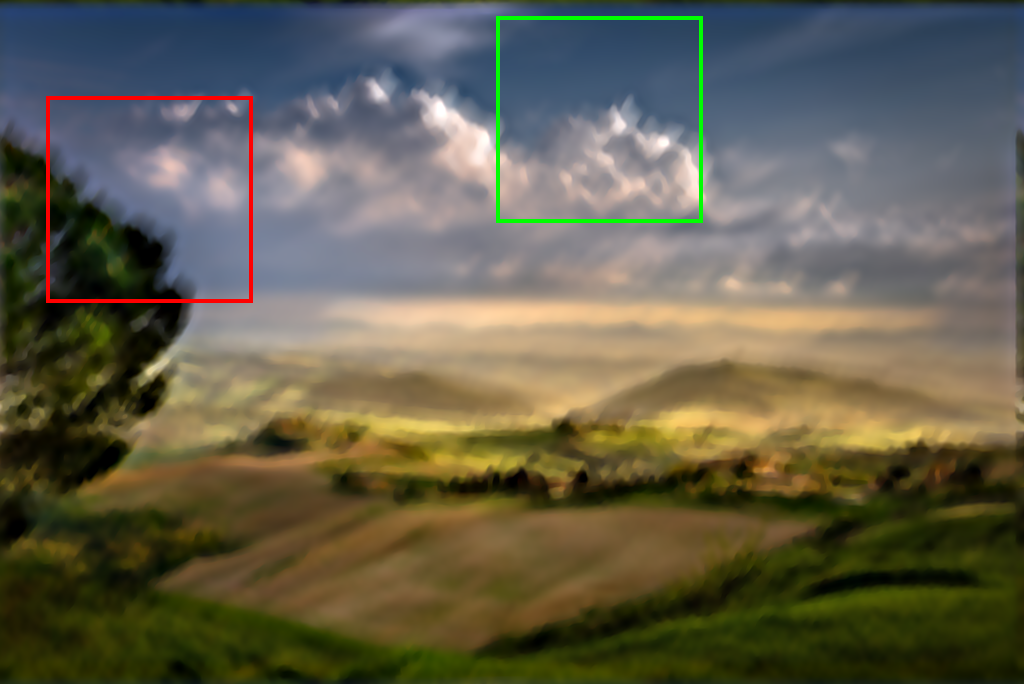}&
			\includegraphics[width=0.105\linewidth,]{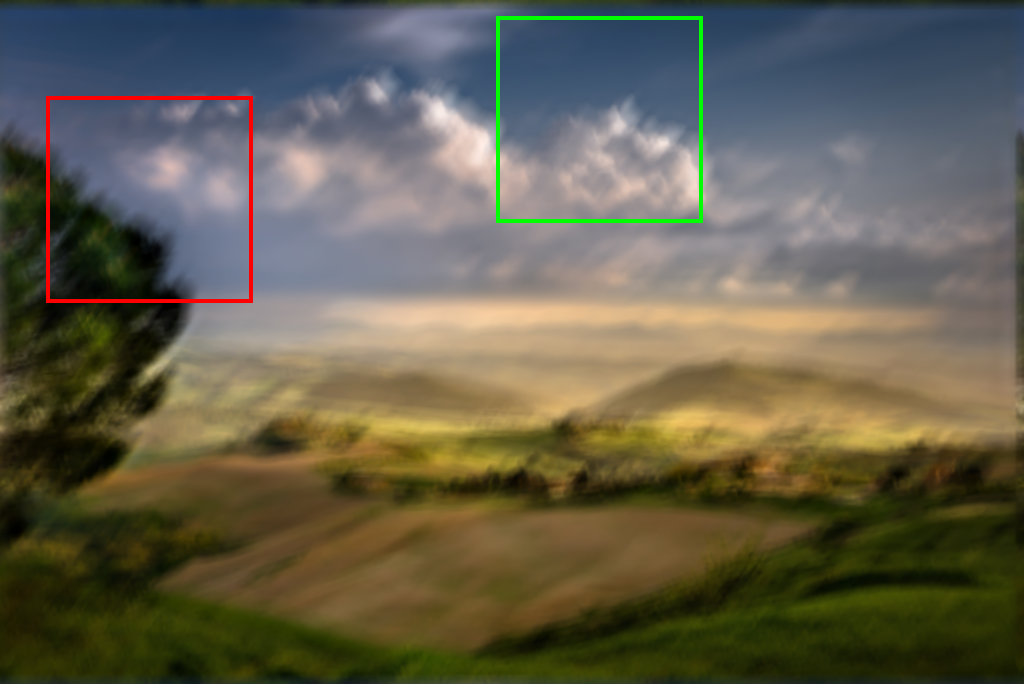}&
			\includegraphics[width=0.105\linewidth,]{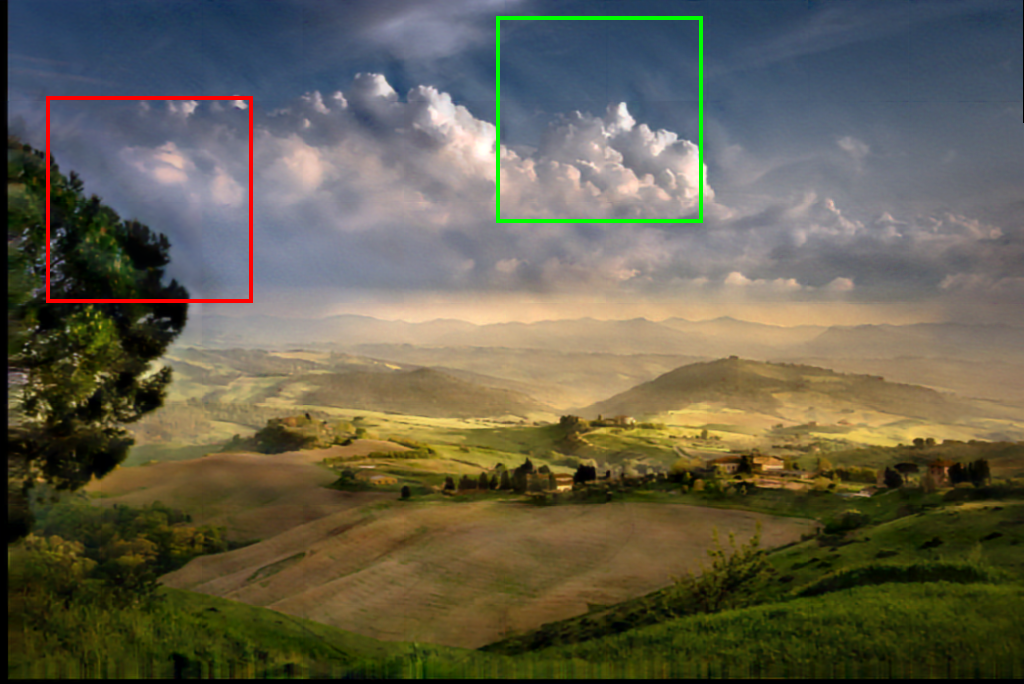}&
			\includegraphics[width=0.105\linewidth,]{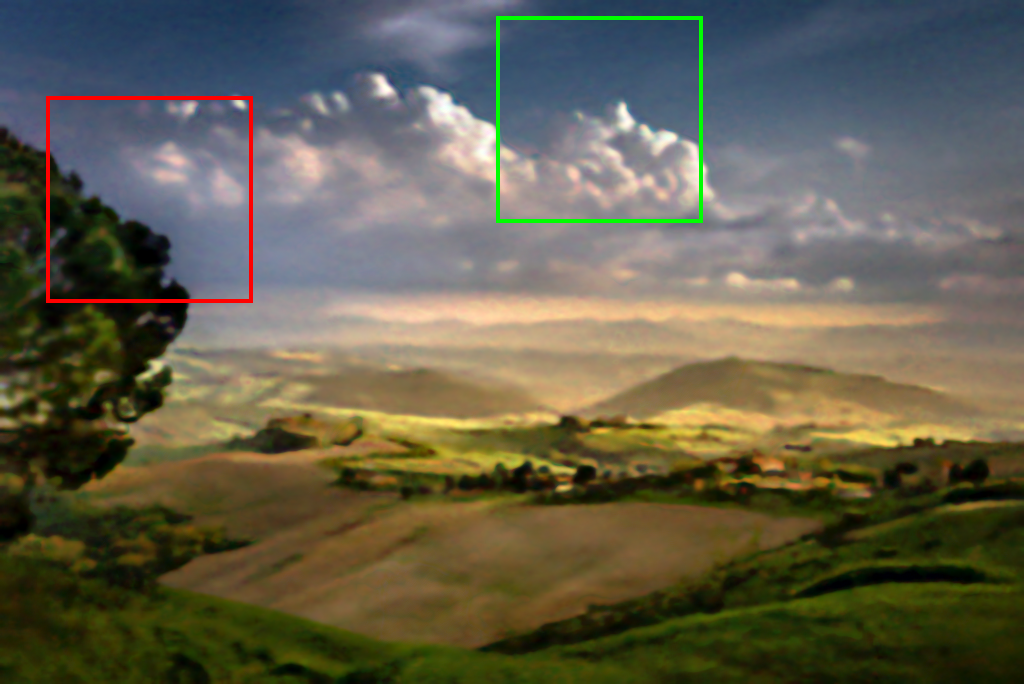}&
			\includegraphics[width=0.105\linewidth,]{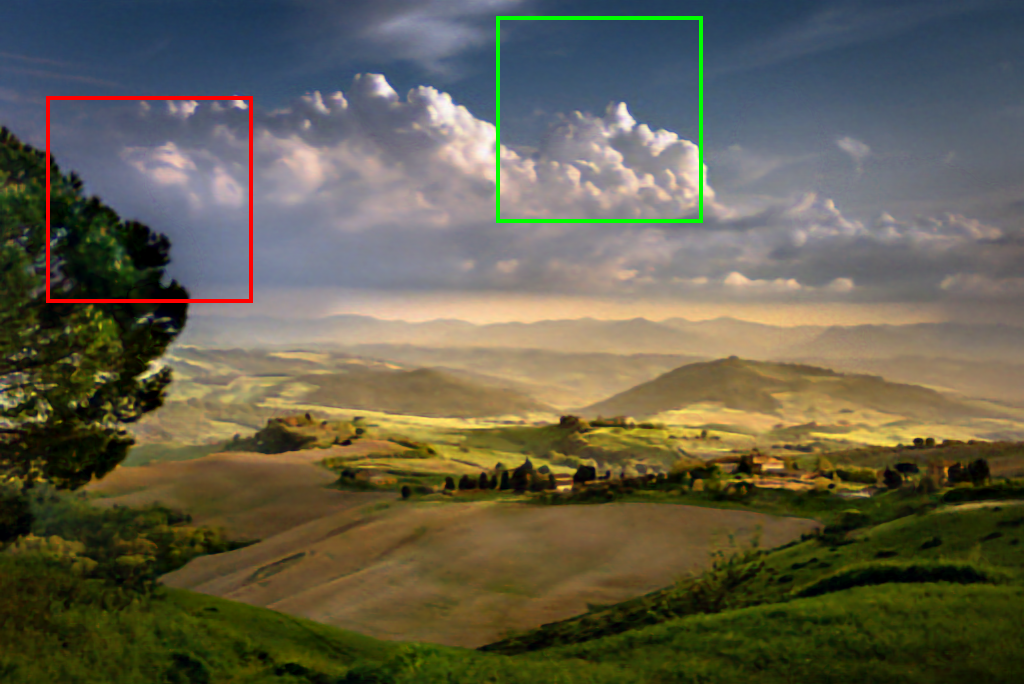}\\
			\includegraphics[width=0.105\linewidth]{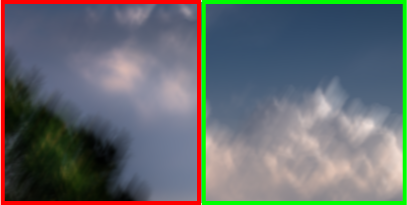}&
			\includegraphics[width=0.105\linewidth]{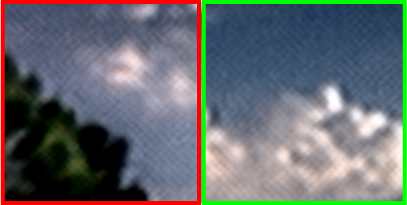}&
			\includegraphics[width=0.105\linewidth]{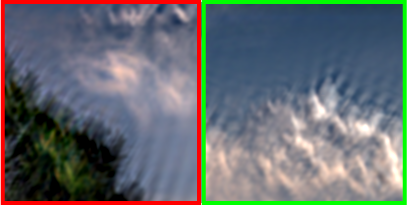}&
			\includegraphics[width=0.105\linewidth]{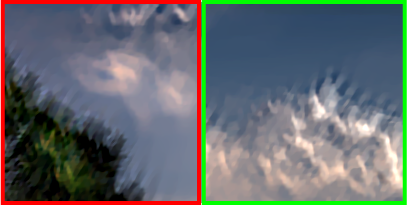}&
			\includegraphics[width=0.105\linewidth]{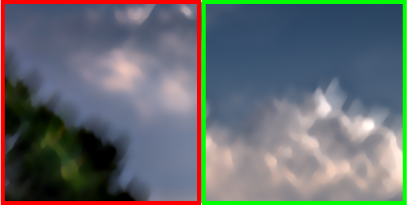}&
			\includegraphics[width=0.105\linewidth]{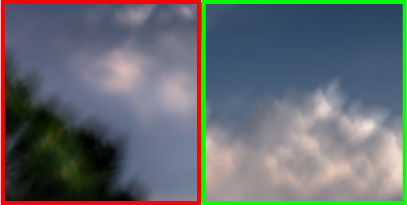}&
			\includegraphics[width=0.105\linewidth]{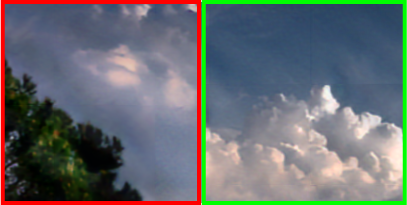}&
			\includegraphics[width=0.105\linewidth]{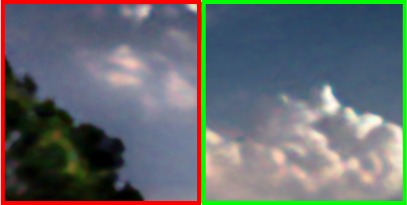}&
			\includegraphics[width=0.105\linewidth]{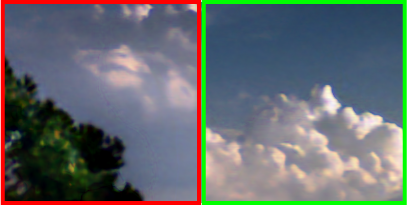}\\
			23.88/0.844&23.50/0.811&21.98/0.781&22.04/0.787&23.86/0.846&23.85/0.843&27.27/0.917&27.86/0.903&\textbf{29.14/0.943}\\
			Blurry&\cite{ren2020neural}&\cite{krishnan2009fast}&\cite{ren2015fast}&\cite{Zhao2013TotalVS}&\cite{ji2011robust}&\cite{vasu2018non}&\cite{Fang_2022_CVPR}&Ours
		\end{tabular}
		\caption{The deblurred results by different methods on the images from the dataset of Levin \etal\cite{levin2009understanding} with the kernel estimated by the method in \cite{cho2009fast} (row 1), and the dataset of Lai \etal \cite{lai2016comparative} with the kernel estimated by the method in \cite{perrone2014total} (row 2). The numerical indexes at the bottom of each image indicate the PSNR/SSIM values. Zoom in for better visualization.}
		\label{Levin}
	\end{figure*}
	
	Tables 2 and 3 list the average PSNR/SSIM values of different methods on the datasets of Levin \etal\cite{levin2009understanding} and Lai \etal \cite{lai2016comparative}. From the tables, we can observe that the proposed method performs best on different blurs and images as compared with the competitive methods. Meanwhile, we observe that the blind method SelfDeblur \cite{ren2020neural} shows unpleasant performance. It is reasonable since SelfDeblur \cite{ren2020neural} does not handle the kernel uncertainty. The reason behind the failure of these non-blind methods, such as \cite{krishnan2009fast, ren2015fast, eboli2020end2end}  should be that they are sensitive to the accuracy of estimated kernels. Among the SBD methods, the proposed method achieves the best quantitative results. The underlying reason is that the proposed method is empowered by a dataset-free DRP, which allows us to faithfully capture the residual as compared to hand-crafted priors \cite{Zhao2013TotalVS, ji2011robust} and data-driven priors \cite{vasu2018non, Fang_2022_CVPR}. 
	
	Figure \ref{reslevin} and Figure \ref{Levin} display the estimated residuals and restored results by different methods.  From Figure \ref{reslevin}, we observe that our method can preserve the fine details of residuals as compared to other methods. In Figure \ref{Levin},  it is not difficult to observe that the deblurred results by our method are much clearer and sharper than others. These observations validate the effectiveness of the proposed method. The visual quality is consistent with the improvement of quantitative performance gain.
	
\begin{figure*}[!t]
		\centering
		\setlength\tabcolsep{1pt}
		\renewcommand{\arraystretch}{0.5} 
		\begin{tabular}{ccccccccc}
			\includegraphics[width=0.105\linewidth]{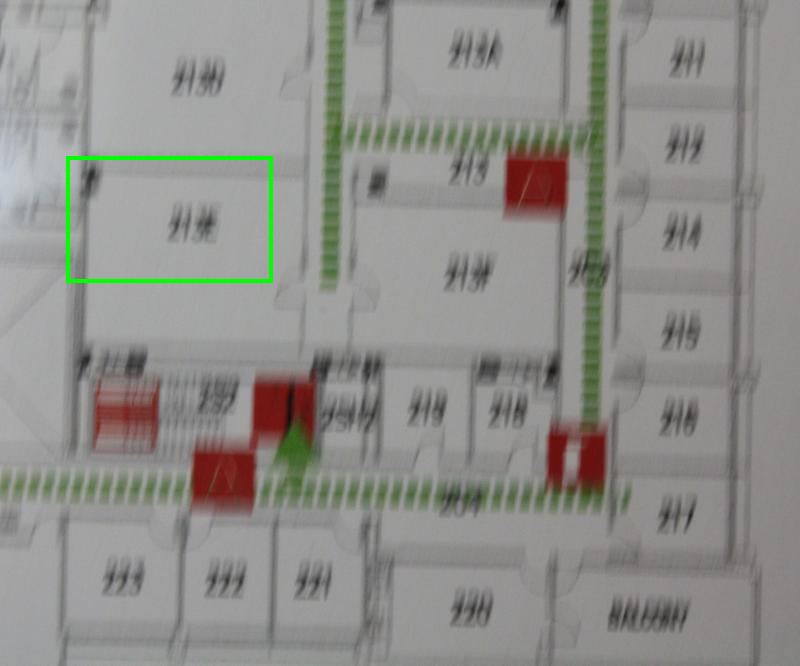}&
			\includegraphics[width=0.105\linewidth]{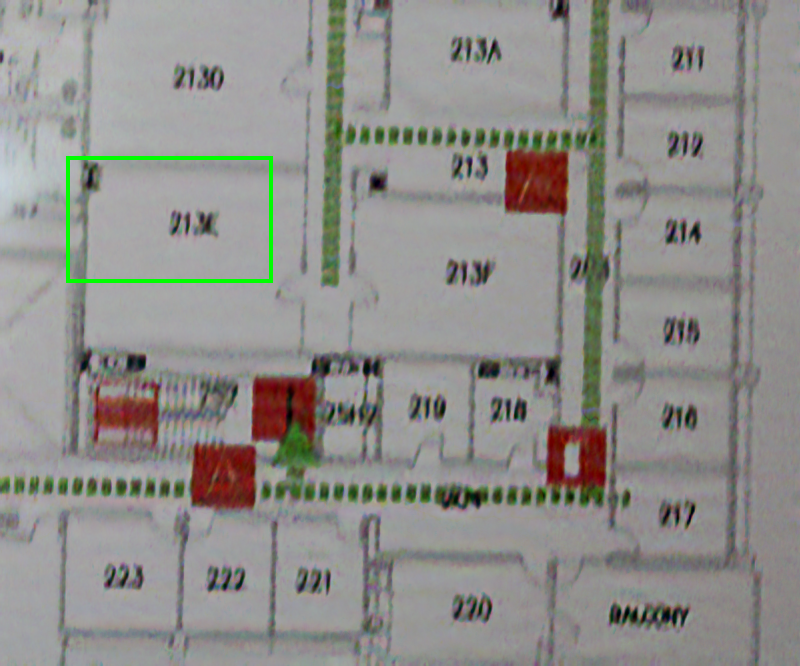}&
			\includegraphics[width=0.105\linewidth]{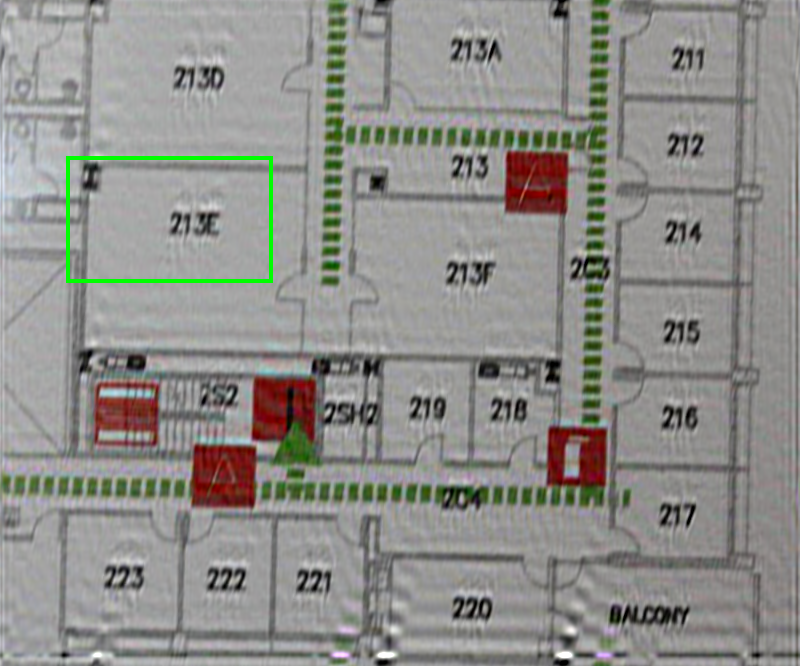}&
			\includegraphics[width=0.105\linewidth]{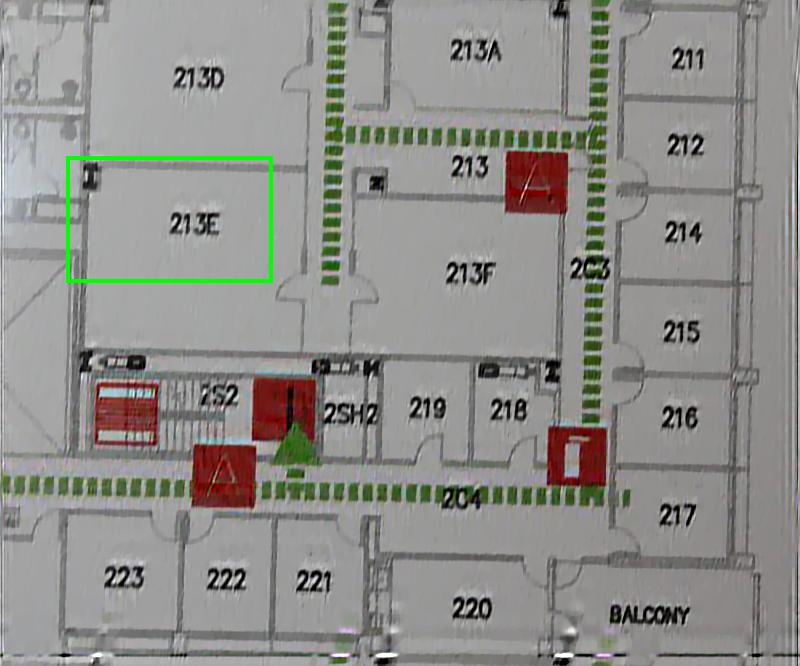}&
			\includegraphics[width=0.105\linewidth]{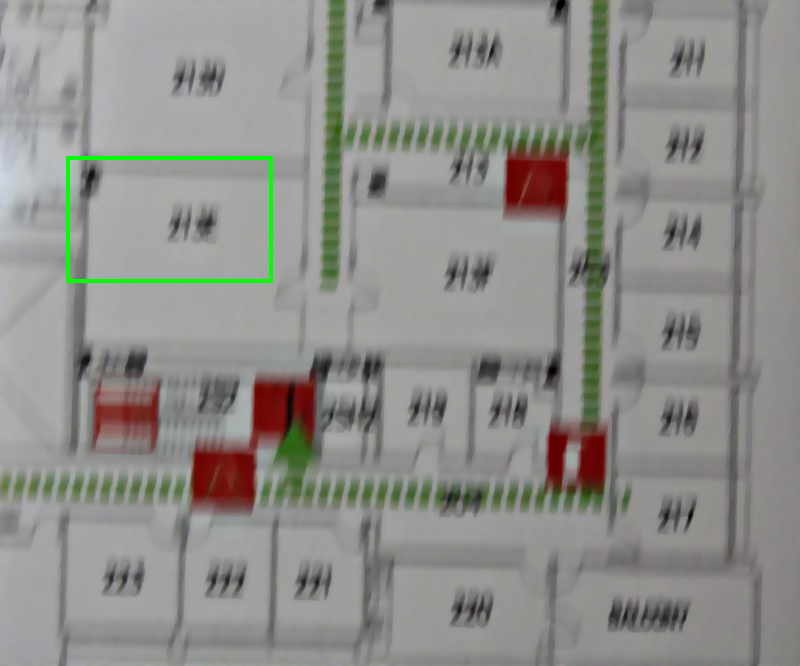}&
			\includegraphics[width=0.105\linewidth]{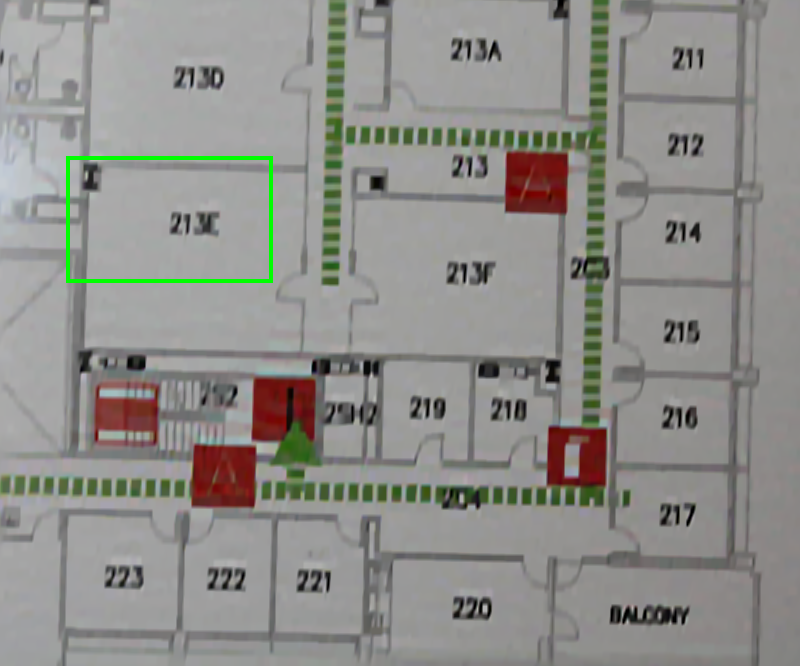}&
			\includegraphics[width=0.105\linewidth]{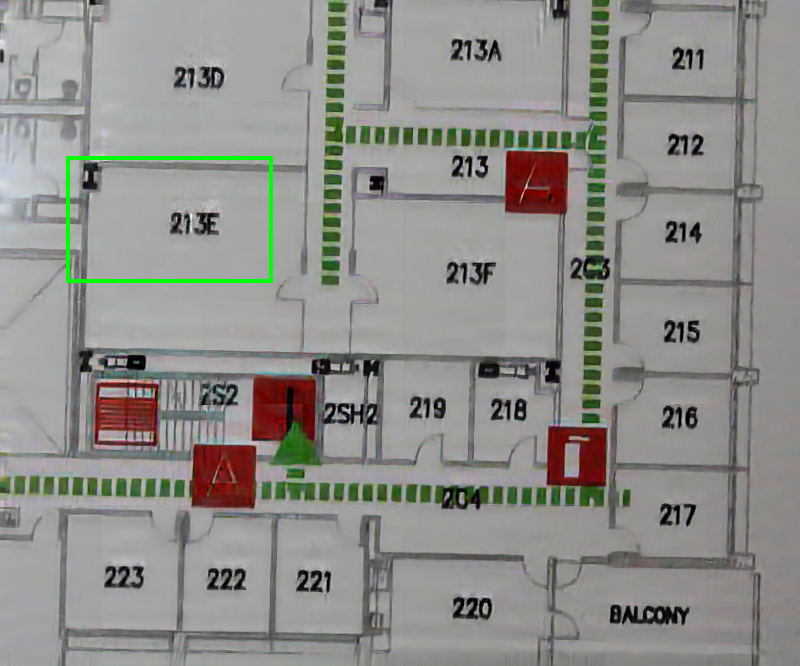}&
			\includegraphics[width=0.105\linewidth]{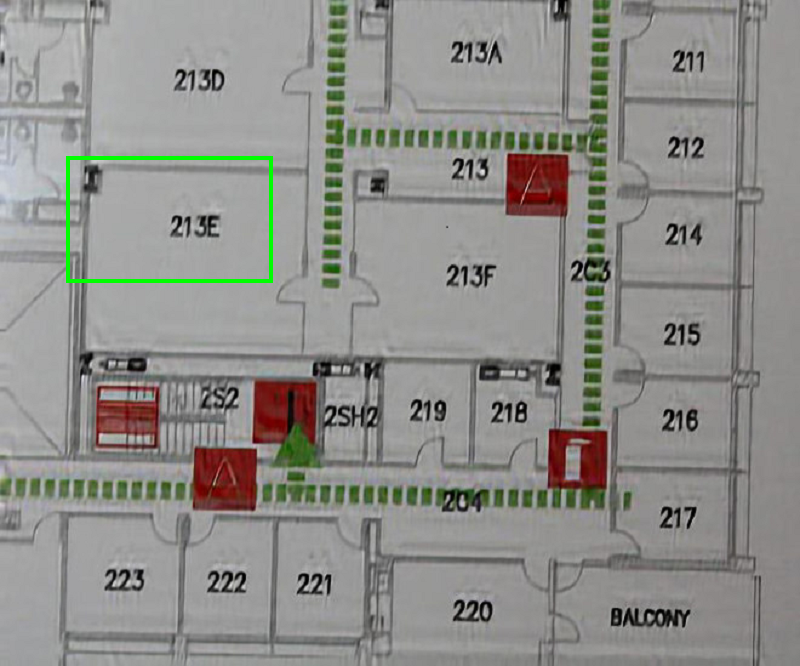}&
			\includegraphics[width=0.105\linewidth]{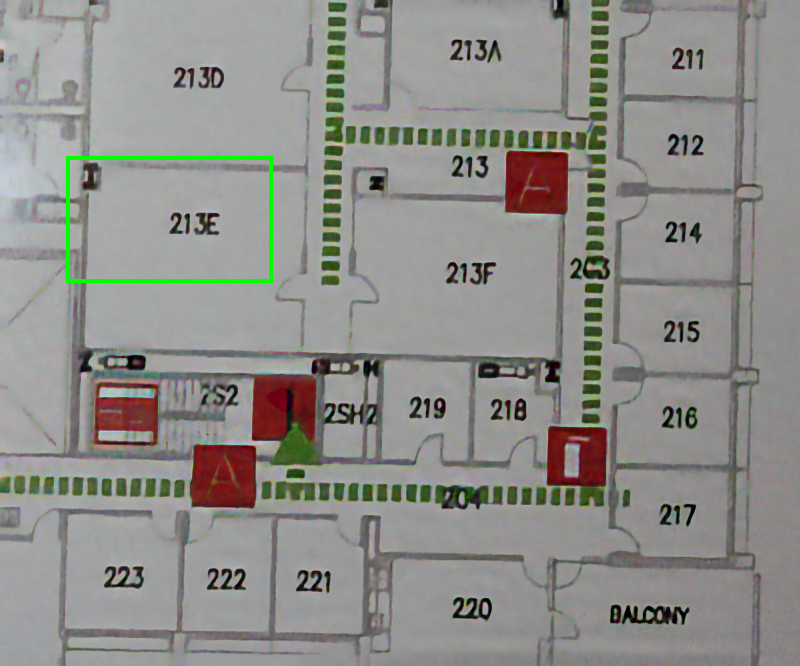}\\
			\includegraphics[width=0.105\linewidth]{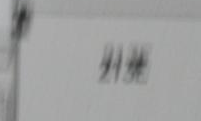}&
			\includegraphics[width=0.105\linewidth]{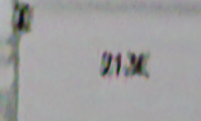}&
			\includegraphics[width=0.105\linewidth]{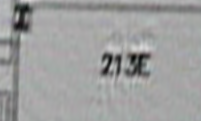}&
			\includegraphics[width=0.105\linewidth]{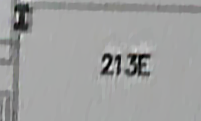}&
			\includegraphics[width=0.105\linewidth]{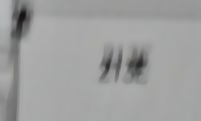}&
			\includegraphics[width=0.105\linewidth]{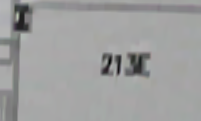}&
			\includegraphics[width=0.105\linewidth]{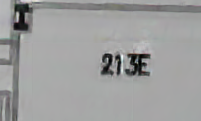}&
			\includegraphics[width=0.105\linewidth]{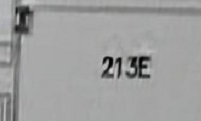}&
			\includegraphics[width=0.105\linewidth]{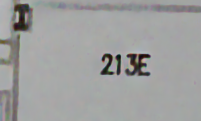}\\
			\quad\\
			\includegraphics[width=0.105\linewidth]{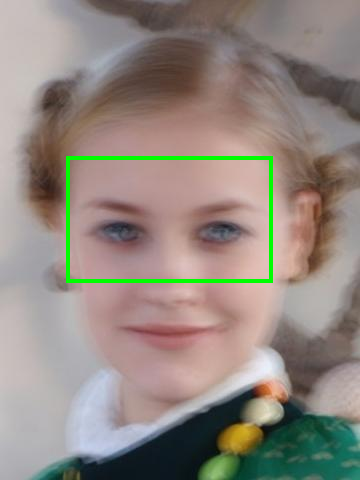}&
			\includegraphics[width=0.105\linewidth]{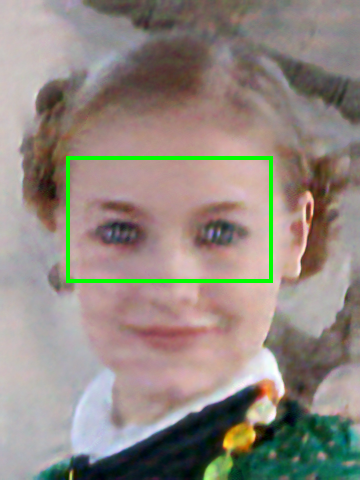}&
			\includegraphics[width=0.105\linewidth]{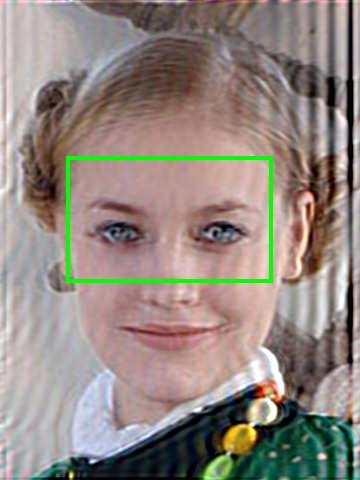}&
			\includegraphics[width=0.105\linewidth]{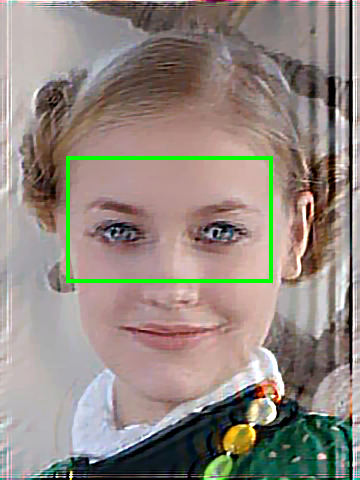}&
			\includegraphics[width=0.105\linewidth]{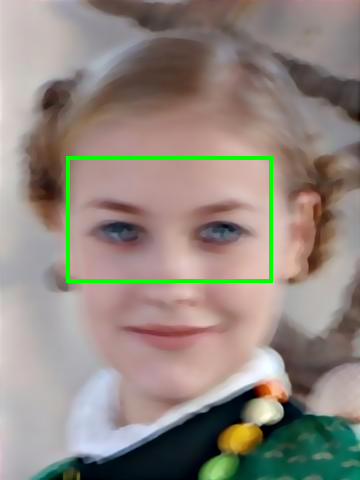}&
			\includegraphics[width=0.105\linewidth]{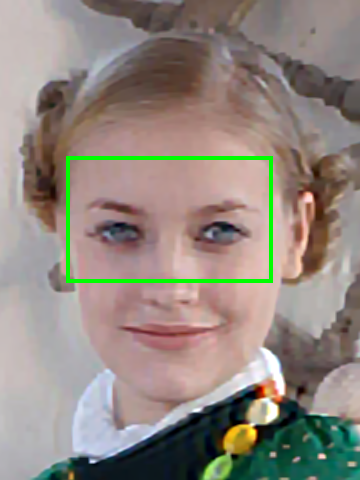}&
			\includegraphics[width=0.105\linewidth]{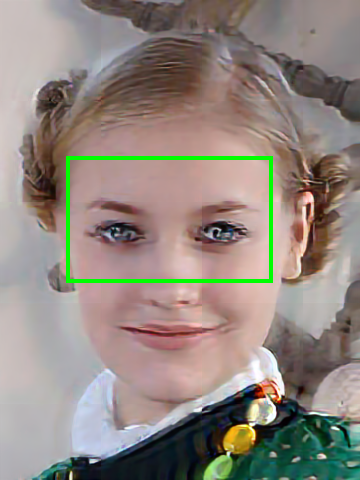}&
			\includegraphics[width=0.105\linewidth]{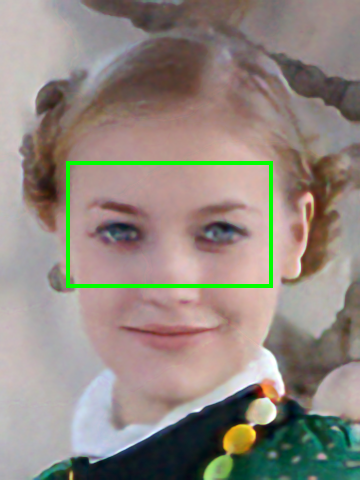}&
			\includegraphics[width=0.105\linewidth]{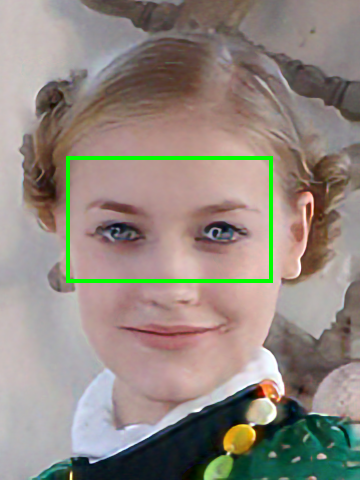}\\
			\includegraphics[width=0.105\linewidth]{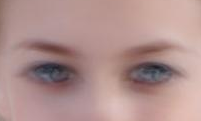}&
			\includegraphics[width=0.105\linewidth]{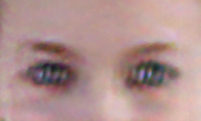}&
			\includegraphics[width=0.105\linewidth]{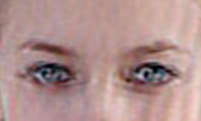}&
			\includegraphics[width=0.105\linewidth]{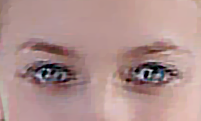}&
			\includegraphics[width=0.105\linewidth]{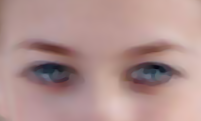}&
			\includegraphics[width=0.105\linewidth]{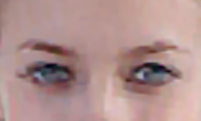}&
			\includegraphics[width=0.105\linewidth]{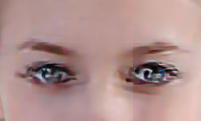}&
			\includegraphics[width=0.105\linewidth]{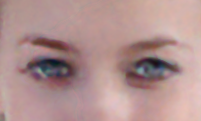}&
			\includegraphics[width=0.105\linewidth]{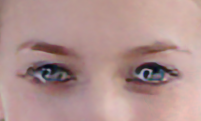}\\
			Blurry&\cite{ren2020neural}&\cite{krishnan2009fast}&\cite{ren2015fast}&\cite{Zhao2013TotalVS}&\cite{ji2011robust}&\cite{vasu2018non} &\cite{Fang_2022_CVPR}&Ours
		\end{tabular}
		\caption{The deblurred results by different methods on the real images from the dataset of Lai \etal\cite{lai2016comparative} with the kernel estimated by the method in \cite{pan2016blind}. Zoom in for better visualization.}
		\label{real}
	\end{figure*}

	Figure \ref{real} displays the deblurred results of two real images from the dataset of Lai \etal\cite{lai2016comparative} with the kernel estimated by the method in \cite{pan2016blind}. We can see that the BD method \cite{ren2020neural} does not deliver faithful deblurring results. NBD methods \cite{krishnan2009fast, ren2015fast} yield significant ringing artifacts in the restored images. Among the SBD methods, model-driven methods  \cite{Zhao2013TotalVS,ji2011robust} perform unpleasantly in deblurring while the data-driven methods \cite{vasu2018non, Fang_2022_CVPR} slightly arouse outliers and artifacts. In comparison, our method, which leverages the power of the DRP, faithfully restores high-quality clear images and robustly handles the artifacts.
	
	\begin{figure*}[!t]
		\centering
		\includegraphics[scale=0.845]{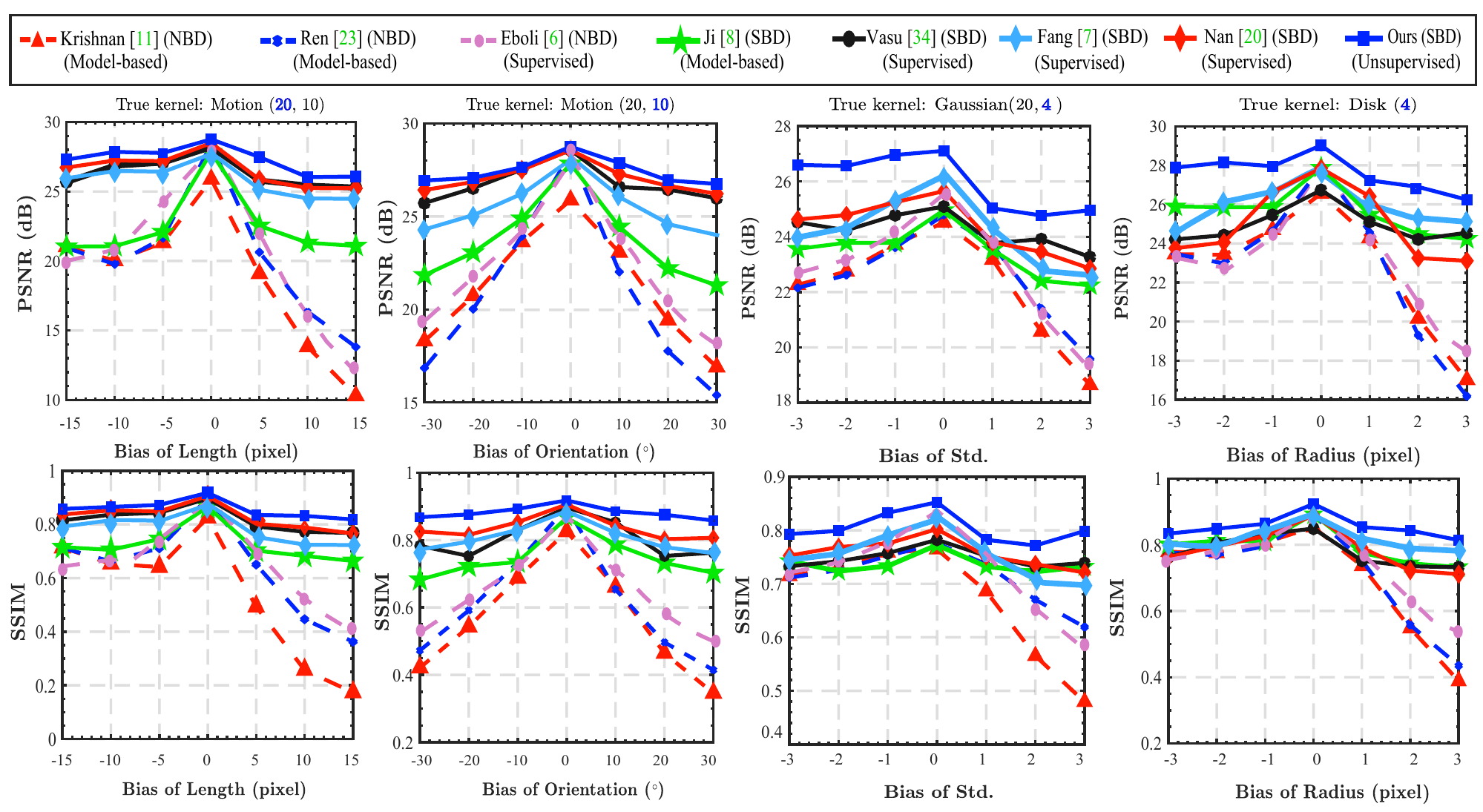}
		\caption{PSNR and SSIM values against bias of inaccurate kernel parameters obtained by different methods. }
		\label{Rob}
	\end{figure*} 
	\subsection{Robustness to Different Types of Kernel Error}
	
	In this section, we discuss the robustness of our method to different types of kernel error. Here we consider  a motion blur with 20 pixels length and $10^\circ$ orientation, denoted by Motion $(20,10)$; a Gaussian blur with size $20\times20$ and $\sigma=4$, denoted by Gaussian $(20,4)$; and a disk-like defocus blur with a radius of 4 pixels, denoted by Disk $(4)$. The test images are generated by blurring the sharp images (see \textbf{supplementary materials}) with the above kernels.
	
	\noindent\textbf{Inaccurate Kernel Setting.} Different types of inaccurate kernels are generated by varying the parameters of different kernels, e.g., the  variance in Gaussian, the orientation and length in Motion, and the radius in Disk. 
	
	\noindent\textbf{Compared Methods.} Five methods are selected as comparisons, including NBD  \cite{krishnan2009fast,ren2015fast} and SBD \cite{vasu2018non,ji2011robust, nan2020deep} methods. In particular, the recent data-driven semi-blind method by Nan and Ji \cite{nan2020deep} is selected as a candidate since this method achieves soft-of-the-art performance and can efficiently estimate the residual on benchmark datasets in a supervised manner.
	
	Figure \ref{Rob} displays the PSNR and SSIM values against the bias of inaccurate kernel parameters. We can observe from the figure that the proposed method delivers pleasant robustness to kernel error and achieves the best quantitative performance in comparison with other methods. Besides, we can see that the NBD methods \cite{krishnan2009fast,ren2015fast} significantly deteriorate as the bias is amplified. The underlying reason is that they are sensitive to kernel error. As for other SBD methods,  the model-driven method \cite{ji2011robust} is inferior to data-driven methods \cite{vasu2018non,nan2020deep} since the hand-crafted prior is generally insufficient in characterizing the complex residual. Data-driven based methods perform well in the case of motion blurs, but fail to continue their impressive performances in the cases of Gaussian and Disk blurs as they are vulnerable to out-of-distribution blurs and images. The reason for our success should be that we utilize  the dataset-free DRP that is expressed by the untrained customized U-Net, which allows us to flexibly adapt to different blurs and images.
	
	\section{Conclusion}
	We proposed a dataset-free DRP for the residual induced by the kernel error depicted by a customized untrained deep neural network, which allowed us to generalize to different blurs and images in real scenarios. With the power of DRP, we then proposed an uncertainty-aware unsupervised image deblurring model  incorporating deep priors and hand-crafted priors, in which all priors work together to deliver a promising performance.  Extensive experiments on different blurs and images showed that the proposed method achieves notable performance as compared with the state-of-the-art methods in terms of image quality and the robustness to different types of kernel error.
	
	\section*{Acknowledgement}
	This research is supported by NSFC (No.12171072, 62131005, 12071069, 12171076), and the National Key Research and Development Program of China (No. 2020YFA0714001, No. 2020YFA0714100).
	
	{\small
		\bibliographystyle{ieee_fullname}
		\bibliography{egbib}

\begin{thebibliography}{10}\itemsep=-1pt

\bibitem{chen2021learning}
Liang Chen, Jiawei Zhang, Jinshan Pan, Songnan Lin, Faming Fang, and Jimmy~S
  Ren.
\newblock Learning a non-blind deblurring network for night blurry images.
\newblock In {\em CVPR}, pages 10542--10550, 2021.

\bibitem{cho2009fast}
Sunghyun Cho and Seungyong Lee.
\newblock Fast motion deblurring.
\newblock In {\em ACM SIGGRAPH}, pages 1--8. 2009.

\bibitem{perrone2014total}
Perrone Daniele and Favaro Paolo.
\newblock Total variation blind deconvolution: The devil is in the details.
\newblock In {\em CVPR}, pages 2909--2916, 2014.

\bibitem{dong2021dwdn}
Jiangxin Dong, Stefan Roth, and Bernt Schiele.
\newblock Dwdn: Deep wiener deconvolution network for non-blind image
  deblurring.
\newblock {\em IEEE TPAMI}, (01):1--1, 2021.

\bibitem{6392274}
Weisheng Dong, Lei Zhang, Guangming Shi, and Xin Li.
\newblock Nonlocally centralized sparse representation for image restoration.
\newblock {\em IEEE TIP}, 22(4):1620--1630, 2013.

\bibitem{eboli2020end2end}
Thomas Eboli, Jian Sun, and Jean Ponce.
\newblock End-to-end interpretable learning of non-blind image deblurring.
\newblock In {\em ECCV}, 2020.

\bibitem{Fang_2022_CVPR}
Yingying Fang, Hao Zhang, Hok~Shing Wong, and Tieyong Zeng.
\newblock A robust non-blind deblurring method using deep denoiser prior.
\newblock In {\em CVPRW}, pages 735--744, June 2022.

\bibitem{ji2011robust}
Hui Ji and Kang Wang.
\newblock Robust image deblurring with an inaccurate blur kernel.
\newblock {\em IEEE TIP}, 21(4):1624--1634, 2011.

\bibitem{jia2007single}
Jiaya Jia.
\newblock Single image motion deblurring using transparency.
\newblock In {\em CVPR}, pages 1--8, 2007.

\bibitem{Adam}
Diederick~P Kingma and Jimmy Ba.
\newblock Adam: A method for stochastic optimization.
\newblock In {\em ICLR}, 2015.

\bibitem{krishnan2009fast}
Dilip Krishnan and Rob Fergus.
\newblock Fast image deconvolution using hyper-laplacian priors.
\newblock {\em NIPS}, 22, 2009.

\bibitem{lai2016comparative}
Wei-Sheng Lai, Jia-Bin Huang, Zhe Hu, Narendra Ahuja, and Ming-Hsuan Yang.
\newblock A comparative study for single image blind deblurring.
\newblock In {\em CVPR}, pages 1701--1709, 2016.

\bibitem{levin2009understanding}
Anat Levin, Yair Weiss, Fredo Durand, and William~T Freeman.
\newblock Understanding and evaluating blind deconvolution algorithms.
\newblock In {\em CVPR}, pages 1964--1971, 2009.

\bibitem{levin2011efficient}
Anat Levin, Yair Weiss, Fredo Durand, and William~T Freeman.
\newblock Efficient marginal likelihood optimization in blind deconvolution.
\newblock In {\em CVPR}, pages 2657--2664, 2011.

\bibitem{li2022supervised}
Ji Li, Yuesong Nan, and Hui Ji.
\newblock Un-supervised learning for blind image deconvolution via monte-carlo
  sampling.
\newblock {\em Inverse Probl}, 38(3):035012, 2022.

\bibitem{xu2013unnatural}
Xu Li, Zheng Shicheng, and Jia Jiaya.
\newblock Unnatural $l_0$ sparse representation for natural image deblurring.
\newblock In {\em CVPR}, pages 1107--1114, 2013.

\bibitem{liu2014blind}
Guangcan Liu, Shiyu Chang, and Yi Ma.
\newblock Blind image deblurring using spectral properties of convolution
  operators.
\newblock {\em IEEE TIP}, 23(12):5047--5056, 2014.

\bibitem{liu2021surface}
Jun Liu, Ming Yan, and Tieyong Zeng.
\newblock Surface-aware blind image deblurring.
\newblock {\em IEEE TPAMI}, 43(3):1041--1055, 2021.

\bibitem{mou2022deep}
Chong Mou, Qian Wang, and Jian Zhang.
\newblock Deep generalized unfolding networks for image restoration.
\newblock In {\em Proceedings of the IEEE/CVF Conference on Computer Vision and
  Pattern Recognition}, pages 17399--17410, 2022.

\bibitem{nan2020deep}
Yuesong Nan and Hui Ji.
\newblock Deep learning for handling kernel/model uncertainty in image
  deconvolution.
\newblock In {\em CVPR}, pages 2388--2397, 2020.

\bibitem{pan2016blind}
Jinshan Pan, Deqing Sun, Hanspeter Pfister, and Ming-Hsuan Yang.
\newblock Blind image deblurring using dark channel prior.
\newblock In {\em CVPR}, pages 1628--1636, 2016.

\bibitem{quan2021nonblind}
Yuhui Quan, Peikang Lin, Yong Xu, Yuesong Nan, and Hui Ji.
\newblock Nonblind image deblurring via deep learning in complex field.
\newblock 2021.

\bibitem{ren2015fast}
Dongwei Ren, Hongzhi Zhang, David Zhang, and Wangmeng Zuo.
\newblock Fast total-variation based image restoration based on derivative
  alternated direction optimization methods.
\newblock {\em Neurocomputing}, 170:201--212, 2015.

\bibitem{ren2020neural}
Dongwei Ren, Kai Zhang, Qilong Wang, Qinghua Hu, and Wangmeng Zuo.
\newblock Neural blind deconvolution using deep priors.
\newblock In {\em CVPR}, pages 3341--3350, 2020.

\bibitem{ren2017partial}
Dongwei Ren, Wangmeng Zuo, David Zhang, Jun Xu, and Lei Zhang.
\newblock Partial deconvolution with inaccurate blur kernel.
\newblock {\em IEEE TIP}, 27(1):511--524, 2017.

\bibitem{ren2019simultaneous}
Dongwei Ren, Wangmeng Zuo, David Zhang, Lei Zhang, and Ming-Hsuan Yang.
\newblock Simultaneous fidelity and regularization learning for image
  restoration.
\newblock {\em IEEE TPAMI}, 43(1):284--299, 2019.

\bibitem{ren2016image}
Wenqi Ren, Xiaochun Cao, Jinshan Pan, Xiaojie Guo, Wangmeng Zuo, and Ming-Hsuan
  Yang.
\newblock Image deblurring via enhanced low-rank prior.
\newblock {\em IEEE TIP}, 25(7):3426--3437, 2016.

\bibitem{fergus2006removing}
Fergus Rob, Singh Barun, Hertzmann Aaron, Roweis~Sam T, and Freeman~William T.
\newblock Removing camera shake from a single photograph.
\newblock In {\em ACM SIGGRAPH}, pages 787--794. 2006.

\bibitem{BP}
Ra{\'u}l Rojas.
\newblock {\em The Backpropagation Algorithm}, pages 149--182.
\newblock Springer Berlin Heidelberg, Berlin, Heidelberg, 1996.

\bibitem{sun2015learning}
Jian Sun, Wenfei Cao, Zongben Xu, and Jean Ponce.
\newblock Learning a convolutional neural network for non-uniform motion blur
  removal.
\newblock In {\em CVPR}, pages 769--777, 2015.

\bibitem{sun2013edge}
Libin Sun, Sunghyun Cho, Jue Wang, and James Hays.
\newblock Edge-based blur kernel estimation using patch priors.
\newblock In {\em IEEE ICCP}, pages 1--8, 2013.

\bibitem{michaeli2014blind}
Michaeli Tomer and Irani Michal.
\newblock Blind deblurring using internal patch recurrence.
\newblock In {\em ECCV}, pages 783--798, 2014.

\bibitem{imaulyanov2018deep}
Dmitry Ulyanov, Andrea Vedaldi, and Victor Lempitsky.
\newblock Deep image prior.
\newblock In {\em CVPR}, pages 9446--9454, 2018.

\bibitem{vasu2018non}
Subeesh Vasu, Venkatesh~Reddy Maligireddy, and AN Rajagopalan.
\newblock Non-blind deblurring: Handling kernel uncertainty with cnns.
\newblock In {\em CVPR}, pages 3272--3281, 2018.

\bibitem{wang2008new}
Yilun Wang, Junfeng Yang, Wotao Yin, and Yin Zhang.
\newblock A new alternating minimization algorithm for total variation image
  reconstruction.
\newblock {\em SIAM J. Imaging Sci.}, 1(3):248--272, 2008.

\bibitem{wang2022general}
Z Wang, X Cun, J Bao, W Zhou, J Liu, and H~Uformer Li.
\newblock A general u-shaped transformer for image restoration.
\newblock In {\em Proceedings of the IEEE/CVF Conference on Computer Vision and
  Pattern Recognition, New Orleans, LA, USA}, pages 19--24, 2022.

\bibitem{xu2010two}
Li Xu and Jiaya Jia.
\newblock Two-phase kernel estimation for robust motion deblurring.
\newblock In {\em ECCV}, pages 157--170, 2010.

\bibitem{yan2017image}
Yanyang Yan, Wenqi Ren, Yuanfang Guo, Rui Wang, and Xiaochun Cao.
\newblock Image deblurring via extreme channels prior.
\newblock In {\em CVPR}, pages 4003--4011, 2017.

\bibitem{zamir2022restormer}
Syed~Waqas Zamir, Aditya Arora, Salman Khan, Munawar Hayat, Fahad~Shahbaz Khan,
  and Ming-Hsuan Yang.
\newblock Restormer: Efficient transformer for high-resolution image
  restoration.
\newblock In {\em Proceedings of the IEEE/CVF Conference on Computer Vision and
  Pattern Recognition}, pages 5728--5739, 2022.

\bibitem{zhang2017learning1}
Jiawei Zhang, Jinshan Pan, Wei-Sheng Lai, Rynson~WH Lau, and Ming-Hsuan Yang.
\newblock Learning fully convolutional networks for iterative non-blind
  deconvolution.
\newblock In {\em CVPR}, pages 3817--3825, 2017.

\bibitem{Zhao2013TotalVS}
Xile Zhao, Wei Wang, Tieyong Zeng, Tingzhu Huang, and Michael~K. Ng.
\newblock Total variation structured total least squares method for image
  restoration.
\newblock {\em SIAM J. Imaging Sci.}, 35, 2013.

\bibitem{zhou2011coded}
Changyin Zhou, Stephen Lin, and Shree~K Nayar.
\newblock Coded aperture pairs for depth from defocus and defocus deblurring.
\newblock {\em IJCV}, 93(1):53--72, 2011.

\bibitem{zuo2016learning}
Wangmeng Zuo, Dongwei Ren, David Zhang, Shuhang Gu, and Lei Zhang.
\newblock Learning iteration-wise generalized shrinkage--thresholding operators
  for blind deconvolution.
\newblock {\em IEEE TIP}, 25(4):1751--1764, 2016.

\end{thebibliography}
	}
	
\end{document}